\documentclass{bytedance_seed}
\usepackage{amsmath,amssymb,mathtools,amsthm}
\usepackage{algorithm,algorithmic}
\usepackage{array,tabularx,float,longtable,colortbl}
\usepackage{needspace}
\usepackage{fontawesome5}
\newcommand{\name}{MoLT}
\newcommand{\attn}{MoLA}

\newcommand{\softmax}{\operatorname{softmax}}

\newcommand{\rope}{\operatorname{RoPE}}

\newcommand{\para}[1]{\paragraph{#1}}
\newtheorem{proposition}{Proposition}
\newtheorem{corollary}{Corollary}
\newcommand{\code}[1]{\texttt{\small #1}}
\hypersetup{pdftitle={Mixture-of-Loops Transformer},
  pdfauthor={Lianghui Zhu et al.}}
\crefname{figure}{Fig.}{Figs.}
\Crefname{figure}{Fig.}{Figs.}
\crefname{subfigure}{Fig.}{Figs.}
\Crefname{subfigure}{Fig.}{Figs.}
\crefname{table}{Table}{Tables}
\Crefname{table}{Table}{Tables}
\crefname{subtable}{Table}{Tables}
\Crefname{subtable}{Table}{Tables}
\crefname{section}{Section}{Sections}
\Crefname{section}{Section}{Sections}
\crefname{subsection}{Section}{Sections}
\Crefname{subsection}{Section}{Sections}
\crefname{subsubsection}{Section}{Sections}
\Crefname{subsubsection}{Section}{Sections}
\crefname{proposition}{Proposition}{Propositions}
\Crefname{proposition}{Proposition}{Propositions}
\crefname{corollary}{Corollary}{Corollaries}
\Crefname{corollary}{Corollary}{Corollaries}
\crefname{equation}{Eq.}{Eqs.}
\Crefname{equation}{Eq.}{Eqs.}
\crefname{appendix}{Appendix}{Appendices}
\Crefname{appendix}{Appendix}{Appendices}
\crefname{subappendix}{Appendix}{Appendices}
\Crefname{subappendix}{Appendix}{Appendices}
\crefname{subsubappendix}{Appendix}{Appendices}
\Crefname{subsubappendix}{Appendix}{Appendices}
\pretocmd{\section}{\Needspace{5\baselineskip}}{}{}
\pretocmd{\subsection}{\Needspace{4\baselineskip}}{}{}

\title{Mixture-of-Loops Transformer}
\author[1,2]{Lianghui Zhu}
\author[2,\dagger]{Shijie Wang}
\author[1]{Xinggang Wang}
\author[2]{Jinze Bai}
\author[3]{Bo Wang}
\author[1]{Xiang Bai}
\affiliation[1]{Huazhong University of Science \& Technology}
\affiliation[2]{ByteDance Seed}
\affiliation[3]{University of Toronto}

\abstract{
\textsf{\textbf{Abstract:}}
Looped Transformers boost expressivity by reusing shared layers across loops, yet at the cost of efficiency.
Pushing their expressivity--efficiency frontier requires rethinking the core mechanism 
connecting adjacent loops: \emph{What should be passed as a Transformer loops?}
We introduce Mixture-of-Loops Transformer (\name{}), whose Mixture-of-Loops Attention (\attn{}) uses keys and values (KV) as memory for selective reuse across depths and loops.
Under loop residual, MoLT-R retains the residual stream and reads local KV across loops for expressivity. 
Under loop KV, MoLT-KV instead restarts from embeddings and reads current-loop local KV with preceding-loop memory KV at earlier positions for efficiency.
MoLT-KV eliminates same-position cross-loop dependence and, paired with loop forcing for parallel training, unlocks single-traversal Mixture-of-Loops (MoL) decoding beyond the training loop horizon.
To balance efficiency and quality, we introduce U-shaped \name{} (U-\name{}), which separates 
memory construction from prediction readout and skips unused decoder traversals without 
changing predictions.
In 400B-token pretraining, U-\name{} outperforms OLMo3 across 700M/1.5B dense and 3B MoE models, achieving $1.7\times$ validation cross-entropy (CE) token efficiency at 700M while improving 3B MoE 10-domain perplexity by 0.75 and 10-task accuracy by 2.1 percentage points.
In 100B-token continued pretraining of Qwen3.5-0.8B-Base, our U-\name{} adaptation lowers CE on all 11 domains over the matched baseline with $1.4\times$ mean token efficiency.
At inference, U-\name{} with MoL-Hybrid-2 matches non-looped OLMo3 latency at 996.6 vs.\ 1,000.6\,ms on S-size ($\sim$113M) models.
}

\newcommand\checknoteformat[2][]{{\small $^{#1}${\checkdatafont\sffamily\bfseries #2}}}
\renewcommand\checkdata[2][]{\addtolist[#1]{#2}{\checkdatalist}{\checkdataformat}{\qquad}}
\newcommand\checknote[2][]{\addtolist[#1]{#2}{\checkdatalist}{\checknoteformat}{\qquad}}
\checknote[\dagger]{Project Lead}
\correspondence{\email{xgwang@hust.edu.cn}}
\checkdata[Code]{\url{https://github.com/hustvl/MoLT}}

\begin{document}
\maketitle
\begingroup
\setlength{\intextsep}{6pt}
\begin{figure}[H]
  \centering
  \includegraphics[width=\textwidth]{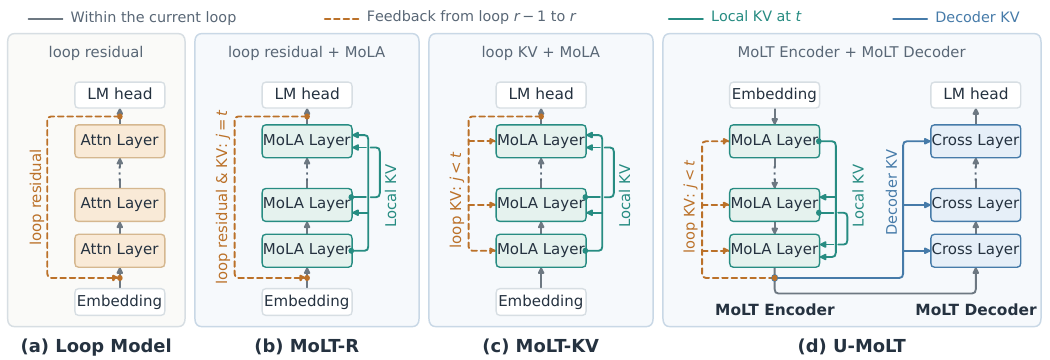}
  \caption{\textbf{MoLT family.}
  (a) A residual-loop reference carries preceding-loop computation through its recurrent state.
  (b) MoLT-R combines loop residual with direct MoLA reads of current-loop and preceding-loop local KV for expressivity.
  (c) MoLT-KV restarts from embeddings and reads preceding-loop memory KV at $j<t$ and current-loop local KV via MoLA for efficiency, enabling single-traversal MoL decoding.
  (d) U-MoLT pairs a loop KV MoLT Encoder with a causal, cross-only MoLT Decoder,
  separating memory construction from prediction readout for balanced efficiency and quality.
  MoLT-R, MoLT-KV, and the U-MoLT encoder show all-layer MoLA under their feedback forms.
  Feed-forward sublayers and normalization are omitted.
  See \cref{sec:discussion} for a detailed comparison.
  }
  \label{fig:teaser}
\end{figure}
\endgroup
\begin{tcolorbox}[colback=white,colframe=seedblue,boxrule=.5pt,
  left=8pt,right=8pt,top=6pt,bottom=6pt]
  \abstractlist
  \par\smallskip
  \checkdatalist
\end{tcolorbox}
\tcbset{reset}
\clearpage
\begingroup
\setlength{\intextsep}{6pt}
\begin{figure}[H]
  \centering
  \includegraphics[width=\textwidth]{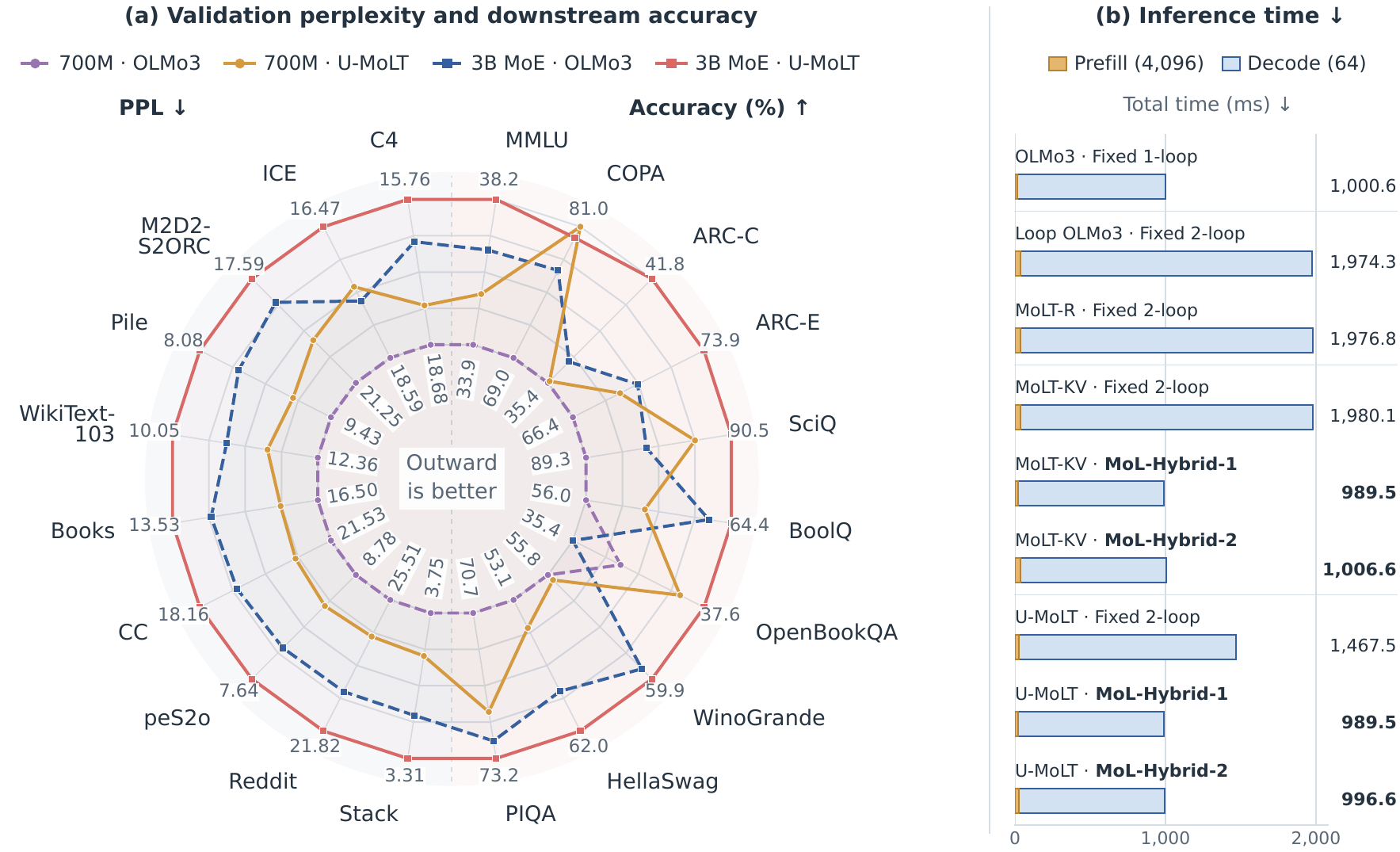}
  \caption{\textbf{Model quality and inference efficiency.}
  (a) Validation perplexity across 10 domains on the left and downstream accuracy across 10 tasks on the right, comparing OLMo3 and U-\name{} at 700M dense and 3B MoE scales after 400B tokens.
  U-\name{} consistently outperforms OLMo3 at both scales.
  Each axis is linearly scaled between its observed extrema, labeled at the inner (worst) and outer (best) rings, so that outward indicates better performance.
  (b) Inference latency in milliseconds of S-size ($\sim$113M) models for 4,096-token prefill and 64 single-token decoding steps.
  Using 1- and 2-loop parallel prefill followed by single-traversal MoL decoding, MoL-Hybrid-1 and MoL-Hybrid-2 reduce fixed 2-loop latency to a level comparable to single-loop OLMo3.}
  \label{fig:results_overview}
\end{figure}
\endgroup
\section{Introduction}
\label{sec:intro}

Model architecture design is fundamentally a tradeoff between expressivity and efficiency.
Recent frontier architectures have largely advanced this tradeoff through sparser computation, linear attention, and richer cross-layer connectivity~\citep{sun2024yoco,yang2025gateddelta,zhu2026moda,kimi2026attnres,deepseekai2026deepseekv41flash,team2026kimik3}.
More recently, looped Transformers have risen to the forefront of architecture design and achieved remarkable success~\citep{dehghani2019universal,geiping2025recurrent,zhu2025ouro,wang2026smelt,chen2026loopscaling,zhao2026unlockingpowerloopspretrained}.
This stems from reusing parameters across repeated loops for stronger expressivity, though at the cost of efficiency.
Building on this paradigm, we pursue next-generation foundation model architectures that combine stronger expressivity with high computational efficiency.
Achieving this requires rethinking the core mechanism connecting adjacent loops: \emph{What should be passed as a Transformer loops?}

The standard answer is loop residual, which passes each loop's output directly into the next loop.
This choice bottlenecks both expressivity and efficiency.
Repeatedly compressing computation into a single residual stream causes information dilution across layers and loops, leading to representational degradation.
Moreover, each token depends on its own preceding-loop output, forcing inference to run every loop sequentially.
Instead of relying solely on a residual stream, our answer is to pass computed keys and values (KV) as memory for selective reuse across depths and loops.

We introduce Mixture-of-Loops Transformer (\name{}), whose Mixture-of-Loops Attention (\attn{}) lets each layer jointly attend to KV from different depths and loops (\cref{fig:teaser}).
Under loop residual, MoLT-R keeps the residual stream across loops for maximum expressivity.
\attn{} directly reads current-loop and preceding-loop local KV at the same token position, mitigating information dilution through selective reuse.
When decoding efficiency is the priority, keeping the residual stream still requires sequential loop execution.
To remove this bottleneck, we ask whether each loop can start afresh and rely purely on KV memory.

This motivates the loop KV paradigm, in which each loop restarts from embeddings.
In MoLT-KV, \attn{} jointly reads memory KV projected from the preceding loop's outputs at strictly earlier positions and current-loop local KV at the current token.
We introduce \emph{loop forcing} to train this paradigm, supervising every loop while training each feedback loop in parallel from preceding-loop memory KV rather than sequentially decoded prefixes.
Because each loop restarts from embeddings and reads memory KV only from earlier positions, a token no longer depends on its own preceding-loop output.
This unlocks Mixture-of-Loops (MoL) decoding, where each new token skips the preceding loops and traverses the shared stack once.
It reads cached prefix memory KV alongside current-loop local KV and appends new memory KV for subsequent tokens, extending looped computation beyond the training loop horizon.
For fast prompt processing, MoL-Hybrid decoding further pairs parallel fixed-loop prefill with MoL decoding.

MoLT-KV brings single-traversal decoding to a plain stack while looping all layers.
To further lower prefill cost and balance efficiency with quality, we introduce U-shaped \name{} (U-\name{}), inspired by encoder-source reuse in YOCO, Universal YOCO, and DeepSeek's CED architecture~\citep{sun2024yoco,sun2026universalyoco,deepseekai2026deepseekv41flash}.
U-\name{} separates memory construction from prediction readout.
It pairs a loop KV MoLT Encoder with a cross-only MoLT Decoder that reads the normalized encoder output through independent per-layer decoder KV projections.
The decoder stays outside the encoder loop and does not depend on decoder states at earlier positions.
At inference, U-\name{} therefore skips decoder traversals at intermediate loops and non-final prompt positions without changing the next-token prediction.

Across scales and architectures, \name{} pushes the expressivity and efficiency frontier of looped models (\cref{fig:results_overview}).
In 400B-token from-scratch pretraining against OLMo3, U-\name{} improves mean validation perplexity and downstream accuracy on 700M dense, 1.5B dense, and 3B mixture-of-experts (MoE) models.
At 700M scale, MoLT-R and U-\name{} raise 10-task accuracy by 2.6 and 2.5 percentage points, and U-\name{} matches OLMo3's final mean validation cross-entropy with $1.7\times$ token efficiency.
On 3B MoE models with 660M active parameters, U-\name{} lowers 10-domain perplexity by 0.75 and raises 10-task accuracy by 2.1 percentage points.
Beyond pretraining from scratch, we validate cross-architecture continued pretraining on Qwen3.5-0.8B-Base, where U-\name{} reduces validation cross-entropy on all 11 domains over the matched 100B-token baseline with $1.4\times$ mean token efficiency.
At inference, MoL-Hybrid decoding eliminates the latency penalty of fixed-loop decoding while preserving prediction quality.
On S-size ($\sim$113M) models with 4,096-token prompts and 64 decoding steps, U-\name{} with MoL-Hybrid-2 runs in 996.6\,ms, matching non-looped OLMo3 at 1,000.6\,ms.

The contributions of this paper are summarized as follows:
  \begin{itemize}
    
    \item We address the expressivity--efficiency bottleneck of looped Transformers and propose a new family of model architectures, \name{}, spanning MoLT-R (loop residual), MoLT-KV (loop KV), and U-\name{}. Through \attn{}, \name{} passes computed KV as memory for selective reuse across depths and loops, while U-\name{} separates memory construction from prediction readout.

    \item We introduce loop-forcing training and MoL decoding. To our knowledge, MoL decoding is the first single-traversal scheme that runs looped Transformers beyond the training loop horizon. By reading cached prefix memory KV alongside current-loop local KV, it halves shared-stack decoding traversals relative to fixed 2-loop decoding. With parallel prefill, MoL-Hybrid decoding cuts fixed 2-loop latency by 32--50\% on S-size MoLT-KV and U-\name{} for 4,096-token prompts and 64 decoding steps, matching non-looped OLMo3 latency.

    \item We provide extensive empirical evidence and comprehensive ablations showing that \name{} consistently outperforms non-looped baselines across scales and architectures. In 400B-token pretraining against OLMo3 on 700M and 1.5B dense and 3B MoE models, U-\name{} reaches $1.7\times$ token efficiency at 700M, while lowering 10-domain perplexity by 0.75 and raising 10-task accuracy by 2.1 percentage points on 3B MoE. In addition, in 100B-token continued pretraining of Qwen3.5-0.8B-Base, our U-\name{} adaptation reduces validation cross-entropy on all 11 domains over the matched baseline with $1.4\times$ mean token efficiency.
    
  \end{itemize}
\section{Related Work}
\label{sec:related}

\para{Looped Transformers}
Sharing weights across depth goes back to Universal Transformers and ALBERT~\citep{dehghani2019universal,lan2020albert}.
Recent work scales looped language models across dense and MoE architectures~\citep{geiping2025recurrent,zhu2025ouro,wang2026smelt,chen2026loopscaling}, varies loop depth through adaptive token routing or elastic compute budgets~\citep{bae2025recursions,jeddi2026loopformer}, stabilizes deep unrolls~\citep{prairie2026parcae,li2026deeploop}, and adapts pretrained models into looped backbones~\citep{kapl2026growing,marchenko2026closing,chen2026trainingfree}.
These approaches primarily carry computation across loops through the residual state.
MoLT instead uses KV as memory for selective reuse across depths and loops.
As summarized in \cref{tab:related}, it spans MoLT-R under loop residual, MoLT-KV under loop KV, and the encoder--decoder variant U-MoLT.

\para{Attention across depth}
Beyond standard residual connections, DenseFormer~\citep{pagliardini2024denseformer} aggregates earlier layer outputs with static weights, whereas Mixture-of-Depths Attention (MoDA)~\citep{zhu2026moda} and Attention Residuals~\citep{kimi2026attnres} use input-dependent weights.
In particular, MoDA jointly reads sequence KV and same-token depth KV within a single pass.
In looped architectures, CoTFormer, Dreamer, and RecurTrace expose intermediate representations across repeated computation~\citep{mohtashami2025cotformer,knupp2026dreamer,wang2026recurtrace}.
In contrast, MoLA unifies cross-loop KV reuse and same-token depth attention within one attention distribution.
Under loop residual, MoLT-R augments current-loop local KV with preceding-loop local KV at the same token position, allowing each layer to directly read intermediate representations from earlier layers in the current loop and all layers of the preceding loop.
Under loop KV, MoLT-KV pairs current-loop local KV with strict-prefix memory KV from the preceding loop, separating same-token depth attention from cross-loop contextual feedback.
Here, \emph{feedback} refers to passing computed representations from one loop into the next.

\para{Prefix feedback and loop forcing}
Caching earlier hidden states across segments begins with same-level recurrence in Transformer-XL~\citep{dai2019transformerxl}, whereas Staircase Attention and Feedback Transformer expose higher-level representations from earlier chunks or a weighted layer mixture as memory for subsequent tokens~\citep{ju2022staircase,fan2020feedback}.
In non-looped Transformers, Layer-Condensed KV Cache (LCKV) shares top-layer KV across condensed layers under a strict-prefix mask~\citep{wu2024lckv}, and Recurrent Transformer attends to layer-output prefix KV alongside own-input KV within each layer~\citep{oncescu2026recurrent}.
Parallel iterative refinement has also been studied for latent thought tokens and hidden-state feedback~\citep{wu2025pccot,zeng2025ponderlm2,wang2026fullbandwidth,cai2026temporal,huang2026latent,duggal2026lattice}, whereas Deep Equilibrium Models solve for fixed points via root finding~\citep{bai2019equilibrium}.
MoLT-KV introduces the loop KV paradigm for looped Transformers, pairing strict-prefix memory KV projected per layer from the preceding loop's output with current-loop local KV at the current token.
Through loop forcing, MoLT supervises next-token prediction at every loop from the initial memory-building pass onward, aligning cross-loop representations for recursive reuse beyond the training loop horizon.
Combined with embedding restart and strict-prefix masking, this interface yields finite-prefix stabilization in token order without contraction assumptions or iterative root finding (\cref{sec:prefix_stabilization}).

\para{KV sharing and efficient looped decoding}
Sharing KV caches reduces memory and computation in deep and looped Transformers.
Cross-Layer Attention shares projected KV across adjacent layers~\citep{brandon2024cla}, and recent looped models reduce KV overhead by updating a shared cache, replaying selected loops, or sharing final-loop KV~\citep{vendrell2026melt,licakova2026kvreplay,prairie2026benice}.
Looped Prediction Transformer (LPT) jointly attends to a layer's first-loop KV cache and a current-loop sliding window, yet still relies on residual feedback and executes all loops sequentially per token~\citep{monea2026sharedmemory}.
Parallel Loop Transformer (PLT) reduces decoding latency by overlapping loops across token positions via shifted states~\citep{wu2025parallelloop}, whereas LoopFold batches multiple loop depths for the same token prefix using predicted loop inputs and a corrector~\citep{song2026loopfold}.
Rather than executing or overlapping multiple loops per token, MoLT-KV and U-MoLT eliminate same-position cross-loop dependence through embedding restart and strict-prefix memory KV, unlocking single-traversal MoL decoding beyond the training loop horizon.
Each new token reads cached prefix memory KV and appends its own memory KV in one traversal.
MoL-Hybrid decoding further combines parallel fixed-loop prefill with this single-traversal decoding update, retaining prompt parallelism alongside reduced decoding cost and a single persistent KV history for the shared stack.

\para{Decoder with encoder-sourced KV}
Separating memory construction from prediction readout reduces both KV storage and prefill computation.
YOCO builds a shared global KV cache in a self-decoder and reuses it across a cross-decoder~\citep{sun2024yoco}, and Universal YOCO loops the self-decoder that constructs this cache~\citep{sun2026universalyoco}.
DeepSeek-V4.1-Flash (CED) applies per-layer projections from the encoder output for global attention while retaining decoder-local sliding-window attention~\citep{deepseekai2026deepseekv41flash}.
LooM repeats decoder layers after the first over a fixed global memory~\citep{yang2026loom}, whereas Looped CED repeats the encoder while still evaluating all encoder rounds per token at inference~\citep{leng2026loopedced}.
U-MoLT avoids both multi-loop decoding and cross-position decoder coupling.
Its loop KV MoLT Encoder enables single-traversal MoL decoding with one encoder--decoder pass per generated token, while its causal, cross-only MoLT Decoder reads the normalized encoder output through independent per-layer decoder KV projections outside the encoder feedback path.
U-MoLT can therefore omit unused decoder traversals at earlier prompt positions and intermediate loops without changing requested predictions.

\begin{table}[t]
\centering
\caption{\textbf{Architectural comparison with related methods.}
Each row lists the loop input, sequence KV, same-token depth KV, decoder design, and decoding traversals per generated token.
A dash indicates an absent component or a non-looped architecture.
In MoLT-KV and U-MoLT, loop~1 reads current-loop sequence KV at $j<t$.
Subsequent feedback loops read memory KV projected per layer from the preceding loop's output at $j<t$, whereas MoL decoding reads cached prefix memory KV projected from earlier tokens' outputs at $j<t$.
Here enc.\ and dec.\ abbreviate encoder and decoder traversals, including self-decoder and cross-decoder stacks.}
\label{tab:related}
\footnotesize
\setlength{\tabcolsep}{3pt}
\renewcommand{\arraystretch}{1.2}
\begin{tabularx}{\linewidth}{@{}
  >{\raggedright\arraybackslash}p{0.215\linewidth}
  >{\raggedright\arraybackslash}p{0.145\linewidth}
  >{\raggedright\arraybackslash}X
  >{\raggedright\arraybackslash}p{0.188\linewidth}
  >{\raggedright\arraybackslash}p{0.132\linewidth}
  >{\raggedright\arraybackslash}p{0.12\linewidth}
@{}}
\toprule
\textbf{Method} & \textbf{Loop input} & \textbf{Sequence KV} & \textbf{Depth KV (at $t$)} & \textbf{Decoder} & \multicolumn{1}{>{\raggedright\arraybackslash}m{0.12\linewidth}@{}}{\textbf{Decode traversals}} \\
\midrule
MoDA~\citep{zhu2026moda}
  & --
  & Current layer ($j\le t$)
  & Earlier layers
  & --
  & 1 \\
Feedback Transformer~\citep{fan2020feedback}
  & --
  & Layer-mixture memory ($j<t$)
  & --
  & --
  & 1 \\
LCKV~\citep{wu2024lckv}
  & --
  & Shared top-layer KV ($j<t$)
  & --
  & --
  & 1 \\
Recurrent Transformer~\citep{oncescu2026recurrent}
  & --
  & Layer-output KV ($j<t$)
  & Own input (shared proj.)
  & --
  & 1 \\
\addlinespace[2.5pt]
LPT~\citep{monea2026sharedmemory}
  & Residual state
  & First-loop + window ($j\le t$)
  & --
  & --
  & $N$ \\
PLT~\citep{wu2025parallelloop}
  & Shifted residual
  & First-loop + gated window
  & --
  & --
  & $N$ (overlapped) \\
\addlinespace[2.5pt]
Universal YOCO~\citep{sun2026universalyoco}
  & Residual state
  & Self-decoder ($j\le t$)
  & --
  & Cross-only (shared KV)
  & $N$ enc.\ + 1 dec. \\
DeepSeek CED~\citep{deepseekai2026deepseekv41flash}
  & --
  & Encoder ($j\le t$)
  & --
  & Per-layer KV + local attn.
  & 1 enc.\ + 1 dec. \\
\midrule
\rowcolor[gray]{0.955}
\textbf{MoLT-R}
  & Residual state
  & Current-loop ($j\le t$)
  & Current-loop \&\newline preceding-loop local KV
  & --
  & $N$ \\
\rowcolor[gray]{0.955}
\textbf{MoLT-KV}
  & Embedding restart
  & Loop-output memory KV
  & Current-loop local KV (incl.\ own)
  & --
  & 1 (MoL) \\
\rowcolor[gray]{0.955}
\textbf{U-MoLT}
  & Embedding restart
  & Loop-output memory KV
  & Current-loop local KV (incl.\ own)
  & Cross-only (per-layer KV)
  & 1 enc.\ + 1 dec.\ (MoL) \\
\bottomrule
\end{tabularx}
\end{table}

\section{Mixture-of-Loops Transformer}
\label{sec:method}

Given tokens $x_0,\ldots,x_{T-1}$, a looped Transformer repeatedly applies a shared network to produce representations for predicting $x_{t+1}$ from $x_{\leq t}$.
Each pass through the shared network is called a \emph{loop}.
Passing computation from one loop into the next is referred to as \emph{cross-loop feedback}, and we call subsequent loops after the first \emph{feedback loops}.
In the residual-feedback Loop Model, each loop passes its final state to the next.
Its intermediate representations are not retained for direct reads in later loops, and each token remains dependent on its own preceding-loop state.

\begin{figure}[!htb]
  \centering
  \includegraphics[width=\textwidth]{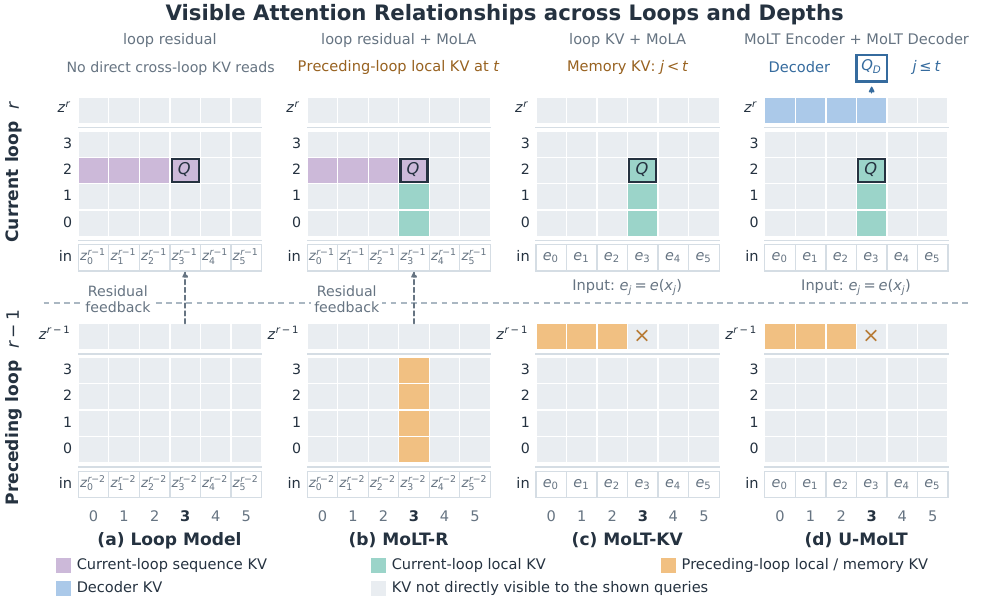}
  \caption{\textbf{Visible attention relationships across loops and depths.}
  Columns index tokens and numbered rows index layers within each loop.
  The highlighted query $Q$ lies at layer $\ell=2$ and token $t=3$ in loop $r>2$, with full-history local KV shown.
  (a) The Loop Model reads current-loop sequence KV at $j\leq t$.
  (b) MoLT-R additionally reads earlier current-loop local KV and all preceding-loop local KV at token $t$.
  (c) MoLT-KV reads memory KV projected from $z^{r-1}$ only at $j<t$, with the cross marking the excluded diagonal, together with current-loop local KV including the layer's own KV at $t$.
  (d) U-MoLT keeps this visibility in the MoLT Encoder and adds a decoder query $Q_D$, whose stack starts from the raw encoder output $X^{r,L_E}$ and reads decoder KV projected from $z^r$ at $j\leq t$ (blue arrow).
  See \cref{sec:discussion} for a detailed comparison.
  }
  \label{fig:attention_visibility}
\end{figure}

We introduce Mixture-of-Loops Transformer (MoLT), which uses KV as memory for selective reuse across depths and loops through Mixture-of-Loops Attention (MoLA) (\cref{fig:teaser,fig:attention_visibility}).
MoLT-R carries the preceding residual state forward and exposes retained local KV, while MoLT-KV restarts from embeddings and receives earlier-loop context through strict-prefix memory KV.
Both are plain MoLT models: their shared loop stack has MoLA at every layer and feeds directly into the prediction head.
U-shaped MoLT (U-MoLT) separates memory construction from prediction readout.
Its MoLT Encoder uses loop KV to build feedback memory across loops, while its separate MoLT Decoder reads the encoder output to predict the next token.
MoLT-R prioritizes expressivity, MoLT-KV brings single-traversal decoding to a plain stack, and U-MoLT balances efficiency and quality through an encoder--decoder split, as compared in \cref{sec:discussion}.
We first define the feedback interfaces and the attention computation they support (\cref{sec:encoder,sec:mola}), separate memory construction from prediction readout (\cref{sec:decoder}), and then present loop-forcing training and MoL decoding (\cref{sec:training,sec:schedules}).

\subsection{Loop Residual and Loop KV}
\label{sec:encoder}
A feedback interface determines what a loop retains and how the next loop receives it.
Each loop initializes a residual stream, applies $L$ blocks of MoLA and feed-forward, and produces an output for prediction and subsequent feedback.
We index loops by $r\geq1$ and layers by $\ell\in\{0,\ldots,L-1\}$.
Layers have distinct parameters within a loop and share those parameters across loops.
We write $X^{r,\ell}$ for the residual stream entering layer $\ell$ and $z^r$ for the normalized loop output.
We write $\operatorname{MoLA}_{\ell}(X,\mathcal S,\mathcal D)$ with three arguments: the layer input $X$, the sequence KV sources $\mathcal S$, and the depth KV sources $\mathcal D$.
MoLA projects its query from $X$ and jointly reads the KV supplied by $\mathcal S$ and $\mathcal D$. These collections contain the selected sources' projected KV pairs, with key normalization and RoPE already applied.
We omit the query indices $(r,\ell,t)$ on the source collections below, and $\sqcup$ concatenates their KV pairs.
MoLA returns the attention update after head concatenation and output projection. FFN denotes the feed-forward sublayer.
For clarity, the layer updates and projections below omit standard block normalization.
The two feedback forms differ in how the retained computation enters this shared network.

\para{Loop residual}
Passing a loop's final state into the next loop provides a direct way to carry its computation forward.
MoLT-R follows this residual feedback path and also retains local KV to make intermediate representations directly readable.
Its sequence KV sources $\mathcal S_X$ are the current layer inputs $X_j^{r,\ell}$ at $j\leq t$, represented by their local KV projections. The subscript $X$ identifies the representation supplying these KV.
Its depth KV sources are $\mathcal D_{\mathrm{cur}}^{<\ell}$ and $\mathcal D_{\mathrm{prev}}$.
Here $\mathcal D_{\mathrm{cur}}^{<\ell}$ contains current-loop local KV projected from earlier-layer inputs $X_t^{r,k}$ with $0\leq k<\ell$, and $\mathcal D_{\mathrm{prev}}$ contains preceding-loop local KV retained from all layers $X_t^{r-1,k}$ with $0\leq k<L$.
We call the reading layer's local KV projected from $X_t^{r,\ell}$ its \emph{own KV}.
Because own KV are already included in $\mathcal S_X$, they are not added to $\mathcal D_{\mathrm{cur}}^{<\ell}$.
The initial state is $e(x_t)$ in loop 1 and $z_t^{r-1}$ thereafter:
\begin{equation}
 \begin{gathered}
 X^{r,0}_t=
 \begin{cases}
 e(x_t),&r=1,\\
 z^{r-1}_t,&r>1,
 \end{cases}
 \qquad \widehat X^{r,\ell}_t
 =X^{r,\ell}_t+\operatorname{MoLA}_{\ell}\!\left(X^{r,\ell}_t,\mathcal S_X,\mathcal D_{\mathrm{cur}}^{<\ell}\sqcup\mathcal D_{\mathrm{prev}}\right),\\[4pt]
 X^{r,\ell+1}_t
 =\widehat X^{r,\ell}_t+\operatorname{FFN}_{\ell}\!\left(\widehat X^{r,\ell}_t\right),
 \qquad z^r=\operatorname{Norm}(X^{r,L}).
 \end{gathered}
 \label{eq:residual_boundary}
\end{equation}
Here $e(x_t)$ denotes the token embedding, $\widehat X^{r,\ell}$ is the residual stream after the attention update, and $\operatorname{Norm}$ is an affine RMSNorm.
For $r=1$, $\mathcal D_{\mathrm{prev}}=\varnothing$ because no preceding-loop local KV exist.
The residual carries the preceding loop's output, while attention selectively reads current-loop and preceding-loop local KV.

\Needspace{7\baselineskip}
\para{Loop KV}
MoLT-KV reuses earlier computation as memory while allowing each loop to start afresh from embeddings.
It reads memory KV projected from the preceding loop's normalized output only at $j<t$.
Embedding restart removes the preceding residual state at $t$, and the strict-prefix mask excludes that position from memory reads.
Together, these rules remove same-position cross-loop dependence while retaining contextual feedback.

Each completed loop supplies the memory representation for the next.
Its sequence KV sources $\mathcal S_{\mathrm{mem}}$ are the preceding-loop outputs $z_j^{r-1}$ at $j<t$, represented by layer $\ell$'s memory KV projections.
In contrast to MoLT-R's $\mathcal D_{\mathrm{cur}}^{<\ell}$, its depth KV sources $\mathcal D_{\mathrm{cur}}$ contain current-loop local KV projected from $X_t^{r,k}$ with $0\leq k\leq\ell$, including own KV at $k=\ell$.
For feedback loops ($r>1$), the forward computation is
\begin{equation}
 \begin{gathered}
 X^{r,0}_t=e(x_t),
 \qquad \widehat X^{r,\ell}_t
 =X^{r,\ell}_t+\operatorname{MoLA}_{\ell}\!\left(X^{r,\ell}_t,\mathcal S_{\mathrm{mem}},\mathcal D_{\mathrm{cur}}\right),\\[4pt]
 X^{r,\ell+1}_t
 =\widehat X^{r,\ell}_t+\operatorname{FFN}_{\ell}\!\left(\widehat X^{r,\ell}_t\right),
 \qquad z^r=\operatorname{Norm}(X^{r,L}).
 \end{gathered}
 \label{eq:boundary}
\end{equation}
The next loop projects its memory KV from $z^r$. 
This interface separates earlier-token context from the current token's computation.
The first loop uses the same MoLA updates with current-loop sequence KV at $j<t$ in place of $\mathcal S_{\mathrm{mem}}$.
Once the prefix memory is ready, the shared network can therefore compute the current token and use its output to extend that memory.

\subsection{Mixture-of-Loops Attention}
\label{sec:mola}
MoLA combines sequence context and same-position depth sources in a single attention distribution, following MoDA's joint sequence--depth normalization~\citep{zhu2026moda}.
For a query at loop $r$, layer $\ell$, and position $t$, we denote the sequence KV sources by $\mathcal S_t^{r,\ell}$ and the depth KV sources by $\mathcal D_t^{r,\ell}$.
The two feedback forms differ in how they construct these two sources.

\para{Sequence KV}
Sequence KV provide context along token positions.
In \emph{loop residual}, $\mathcal S_t^{r,\ell}=\mathcal S_X$ contains current-loop sequence KV from the reading layer at $j\leq t$, projected from its input $X_j^{r,\ell}$:
\begin{equation}
 k^{r,\ell}_{j,\mathrm{loc}}=\rope_j\!\left(\operatorname{Norm}(X^{r,\ell}_jW^\ell_{K,\mathrm{loc}})\right),\qquad
 v^{r,\ell}_{j,\mathrm{loc}}=X^{r,\ell}_jW^\ell_{V,\mathrm{loc}}.
 \label{eq:local}
\end{equation}
Head indices are suppressed throughout.
Each query or key projection uses a separate $\operatorname{Norm}$, applied per head, followed by positional rotation $\rope_j$.

In \emph{loop KV}, sequence context instead comes from the preceding loop's output $z^{r-1}$.
For feedback loops ($r>1$), the reading layer projects memory KV as
\begin{equation}
 k^{r,\ell}_{j,\mathrm{mem}}=\rope_j\!\left(\operatorname{Norm}(z^{r-1}_jW^\ell_{K,\mathrm{mem}})\right),\qquad
 v^{r,\ell}_{j,\mathrm{mem}}=z^{r-1}_jW^\ell_{V,\mathrm{mem}}.
 \label{eq:memory}
\end{equation}
Its sequence sources $\mathcal S_t^{r,\ell}=\mathcal S_{\mathrm{mem}}$ contain these memory KV only at $j<t$, excluding the preceding-loop state at the query position.
The first loop uses current-loop sequence KV at $j<t$ because preceding-loop memory is unavailable, while retaining the same joint sequence--depth attention.

\para{Depth KV}
Depth KV make intermediate representations at the current token directly readable.
Both feedback forms use the local projections in \cref{eq:local} at $j=t$ to produce these sources.
A layer that supplies local KV is a \emph{writer}, and a layer that reads them is a \emph{reader}.
Each local KV pair is available before its writer's attention update and can be retained for later readers.

In \emph{loop residual}, $\mathcal D_t^{r,\ell}=\mathcal D_{\mathrm{cur}}^{<\ell}\sqcup\mathcal D_{\mathrm{prev}}$ combines the selected current-loop and preceding-loop local KV at $t$.
Unlike loop KV's memory KV, these preceding-loop local KV retain their original projections without being reprojected from the loop output.
Preceding-loop writers may be deeper than the current reader because that loop has already finished.
Own KV are already included in $\mathcal S_t^{r,\ell}$ and are not added again.
In the first loop, $\mathcal D_{\mathrm{prev}}=\varnothing$.

In \emph{loop KV}, $\mathcal D_t^{r,\ell}=\mathcal D_{\mathrm{cur}}$ contains only local KV produced within the current loop.
The full-history default includes all earlier layers at $t$ together with the current layer's own KV.
Own KV supply the current-token input excluded from the strict-prefix sequence sources.
Reading these depth sources requires neither the current layer's attention output nor the preceding-loop state $z^{r-1}_t$.

\para{Joint attention}
Both feedback forms project the query from the current layer's input and jointly read $\mathcal S_t^{r,\ell}$ and $\mathcal D_t^{r,\ell}$ with one attention distribution:
\begin{equation}
 \begin{aligned}
 q^{r,\ell}_t=\rope_t\!\left(\operatorname{Norm}(X^{r,\ell}_tW^\ell_Q)\right),\quad%
 o^{r,\ell}_t=
 \frac{\sum_{(k,v)\in\mathcal S_t^{r,\ell}\sqcup\mathcal D_t^{r,\ell}}
                 \exp(s(q^{r,\ell}_t,k))v}
      {\sum_{(k,v)\in\mathcal S_t^{r,\ell}\sqcup\mathcal D_t^{r,\ell}}
                 \exp(s(q^{r,\ell}_t,k))}.
 \end{aligned}
 \label{eq:joint}
\end{equation}
Here $s(q,k)=h^{-1/2}q^\top k$, where $h$ is the head width.
The query therefore determines how much each source contributes, with no separate learned gate over loops.
Concatenating the head outputs and applying the output projection gives the MoLA term in \cref{eq:residual_boundary,eq:boundary}.
Local KV remain available until all their readers finish.

The shared softmax makes the relative logit scales of sequence and depth sources relevant to their weights.
We apply affine RMS QK normalization independently across the dimensions of each query and key head~\citep{henry2020qknorm,zhang2019rmsnorm}, followed by RoPE~\citep{su2024roformer}.
Stored depth keys retain their writer's normalization and rotation, and readers reuse them without further normalization. %

\subsection{U-MoLT: Separating Memory Construction and Prediction Readout}
\label{sec:decoder}
\begin{figure}[!t]
  \centering
  \includegraphics[width=\textwidth]{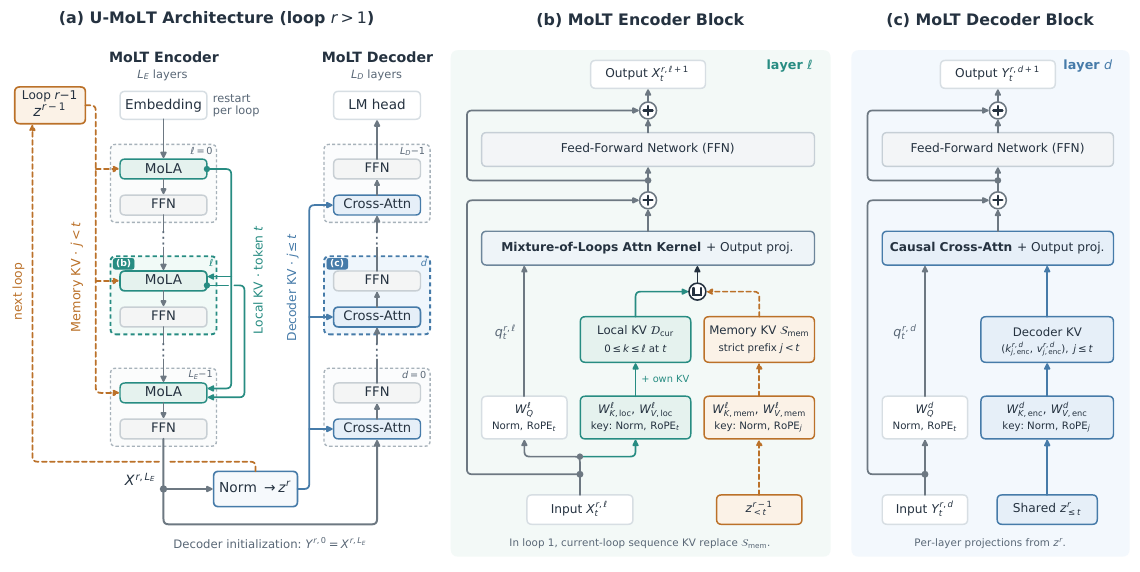}
  \caption{\textbf{U-MoLT separates memory construction from prediction readout.}
  (a) U-MoLT architecture expanding \cref{fig:teaser}(d) for feedback loop $r>1$ with all-layer MoLA and full-history local KV.
  The downward MoLT Encoder restarts from token embeddings, while the upward, cross-only MoLT Decoder starts from the raw encoder output $X^{r,L_E}$.
  The normalized encoder output $z^r$ supplies memory KV for the next loop along orange dashed edges and decoder KV along blue edges, while teal edges carry current-loop local KV at token $t$.
  (b) MoLT Encoder block at layer $\ell$.
  It concatenates current-loop local KV $\mathcal{D}_{\mathrm{cur}}$ from layers $0\leq k\leq\ell$ at token $t$ with strict-prefix memory KV $\mathcal{S}_{\mathrm{mem}}$ projected from $z^{r-1}_{<t}$ for $j<t$ via $\sqcup$ in the Mixture-of-Loops Attention kernel.
  In loop 1, current-loop sequence KV replace $\mathcal{S}_{\mathrm{mem}}$.
  (c) MoLT Decoder block at layer $d$.
  It attends causally to per-layer decoder KV $(k^{r,d}_{j,\mathrm{enc}},v^{r,d}_{j,\mathrm{enc}})$ for $j\leq t$, projected from the normalized encoder output $z^r_{\leq t}$.
  Block normalization is omitted.}
  \label{fig:architecture}
\end{figure}

In plain MoLT-KV, the shared stack constructs feedback memory and feeds its output directly to the prediction head.
Adding layers to this stack for prediction would also deepen memory construction in every loop.
U-MoLT separates these roles by using the loop KV stack as a \emph{MoLT Encoder} and adding a causal, cross-only \emph{MoLT Decoder} for prediction, following YOCO's self-decoder/cross-decoder organization~\citep{sun2024yoco}.
The encoder has $L_E$ MoLA layers and the decoder has $L_D$ cross-attention layers.
The decoder reads the encoder output, while feedback remains determined by the encoder alone (\cref{fig:architecture}).
Prediction readout consists of the prediction head in plain MoLT and the decoder followed by the prediction head in U-MoLT.

The decoder starts from the raw encoder output $Y^{r,0}=X^{r,L_E}$.
Its attention reads decoder KV projected from the normalized encoder output $z^r=\operatorname{Norm}(X^{r,L_E})$.
For layers $d\in\{0,\ldots,L_D-1\}$ with residual input $Y^{r,d}$, the projections are
\begin{equation}
 q^{r,d}_t=\rope_t\!\left(\operatorname{Norm}(Y^{r,d}_tW^d_Q)\right),\quad
 k^{r,d}_{j,\mathrm{enc}}=\rope_j\!\left(\operatorname{Norm}(z^r_jW^d_{K,\mathrm{enc}})\right),\quad
 v^{r,d}_{j,\mathrm{enc}}=z^r_jW^d_{V,\mathrm{enc}}.
 \label{eq:decoder}
\end{equation}
Decoder layer $d$ reads the decoder KV $\{(k^{r,d}_{j,\mathrm{enc}},v^{r,d}_{j,\mathrm{enc}}):j\leq t\}$ and completes its residual and feed-forward updates to produce $Y^{r,d+1}_t$.
All decoder layers share the normalized encoder output $z^r$, while keeping separate decoder KV projections and caches.
The prediction head therefore receives the raw loop output in plain MoLT or the final decoder output in U-MoLT:
\begin{equation}
 p_\theta^r(\cdot\mid x_{\leq t})
 =\softmax\!\left(\operatorname{Norm}(H^r_t)W_{\mathrm{vocab}}\right),\qquad
 H^r=
 \begin{cases}
 X^{r,L},&\text{plain MoLT},\\
 Y^{r,L_D},&\text{U-MoLT}.
 \end{cases}
 \label{eq:forms}
\end{equation}
Here $W_{\mathrm{vocab}}$ is the vocabulary projection.

Keeping the cross-only decoder outside the feedback path allows decoder traversals to be scheduled independently of encoder recurrence.
At inference, U-MoLT can omit intermediate-loop decoder traversals and run the decoder only at the final prompt position without changing the next-token prediction (\cref{sec:cost}).
This separation also keeps decoder updates from altering the normalized encoder output read by later tokens.

\subsection{Loop Forcing: Mixture-of-Loops Training}
\label{sec:training}
We train the MoLT family by supervising next-token predictions at successive loops of the shared network.
Under loop KV, this multi-loop supervision trains the cross-token feedback recurrence in parallel across token positions via loop forcing.

\para{Multi-loop objective}
All loops share the same ground-truth token sequence and next-token targets.
For a general loop count $N$, we combine the per-loop cross-entropy losses as
\begin{equation}
 \mathcal L_r=-\frac{1}{|\Omega|}
       \sum_{t\in\Omega}\log p_\theta^r(x_{t+1}\mid x_{\leq t}),\qquad
 \mathcal L=\sum_{r=1}^{N}\alpha_r\mathcal L_r,\qquad
 \alpha_r=\frac{w_r}{\sum_{k=1}^{N}w_k}.
 \label{eq:loss}
\end{equation}
Here $\Omega$ contains valid prediction positions, $p_\theta^r$ is the model's next-token distribution at loop $r$, and the normalized weights $\alpha_r$ set each loop's contribution.
In U-MoLT, each supervised loop runs the decoder and prediction head to compute its prediction loss.
Each supervised loop trains its prediction path, and we compare loop-loss weights in \cref{sec:components}.

\para{Loop forcing}
For MoLT-KV and U-MoLT, applying \cref{eq:loss} with positive weight on every loop defines \emph{loop forcing}.
Analogous to teacher forcing, each feedback loop trains in parallel across tokens from the preceding loop's strict-prefix memory KV rather than from sequentially decoded prefixes, reducing training from $T$ sequential steps to $N$ parallel passes.
Allowing gradients through cross-loop feedback additionally lets later-loop losses refine the earlier computation that produces the feedback sources, while per-loop supervision remains applicable when cross-loop gradients are detached.
After 2-loop training, single-traversal decoding achieves similar mean continuation CE to fixed 2-loop decoding on the evaluated windows (\cref{sec:execution_results}).
Three design choices help support this shift to recursively generated prefix memory KV (\cref{sec:loop_forcing}).
Shared-readout supervision encourages the supervised loop outputs $z^1$ and $z^2$ to align.
Embedding restart anchors each token's residual stream to $e(x_t)$, while output and QK normalization keep activations and attention logits bounded.

\subsection{Mixture-of-Loops Decoding}
\label{sec:schedules}
We consider three decoding schedules: fixed-loop decoding, Mixture-of-Loops (MoL) decoding, and MoL-Hybrid decoding.
All MoLT variants support fixed-loop decoding, which preserves the chosen finite-loop computation.
MoL decoding combines current-loop local KV with cached prefix memory KV and skips the preceding loops at each new token.
In MoLT-KV and U-MoLT, embedding restart and strict-prefix loop KV feedback remove same-position cross-loop dependencies, enabling one traversal per decoding step.
Both MoL decoding and MoL-Hybrid decoding use this single-traversal update, differing in how they initialize prompt memory KV.
A traversal executes the full Transformer backbone in plain MoLT or the encoder in U-MoLT once, over either the prompt or one token position, while a decoder traversal executes the MoLT Decoder once over the requested positions.
Prefill supplies the first continuation prediction, and each decoding step consumes one token to predict the next.

\para{Fixed-loop decoding}
Fixed-loop prefill executes $N$ traversals of the shared network, processing prompt positions in parallel within each loop. Fixed 2-loop decoding sets $N=2$.
Each loop retains its own prefix sequence KV cache.
During decoding, each new input token passes through all $N$ loops, reading and extending their respective caches before computing the final-loop prediction.
MoLT-R also passes the current token's normalized output and selected local KV from each loop to the next.
These same-position sources require the token's loops to execute in order.
This preserves the chosen finite-loop computation and requires $N$ network traversals per decoding step.

\para{Mixture-of-Loops decoding}
Applied from empty caches, MoL decoding processes the prompt one token at a time.
At each position, one traversal starts from $e(x_t)$, jointly reads completed prefix memory KV at $j<t$ and current-loop local KV through MoLA, and produces a new output.
Memory KV projected from the normalized output are appended to the cache for later tokens.
Decoding uses the same update, so both phases require one traversal per token.

\para{MoL-Hybrid decoding}
MoL-Hybrid decoding combines fixed-loop parallel prefill with the MoL decoding update.
MoL-Hybrid-1 and MoL-Hybrid-2 use 1-loop and 2-loop prefill, respectively, initializing prompt memory KV by projecting $z^1$ and $z^2$.
In U-MoLT, decoder KV come from the final prefill loop.
The prediction at the final prefill position supplies the distribution for the first continuation token.
Each decoding step uses one traversal and extends the retained caches, including memory KV projected from each new output.
The history therefore combines finite-loop prompt caches with KV recursively appended during decoding, trading exact equivalence to an empty-cache sweep for parallel prompt processing.

\para{Key properties and inference costs}
Because embedding restart and strict-prefix memory reads at $j<t$ remove same-position cross-loop dependence, every feedback loop reduces to a strictly triangular recurrence across token positions.
Position $0$ has an empty prefix and stabilizes by loop~2 from any initial state, and each subsequent position stabilizes one loop after its prefix (\cref{sec:prefix_stabilization}).
Empty-cache MoL decoding therefore reaches the unique stabilized solution matched by every unroll of depth $r\geq t+2$.
After a length-$P$ prompt, MoL-Hybrid decoding preserves the same cross-token recursion over continuation positions $m=t-P\geq 0$, matching any multi-loop unroll that holds the prompt fixed and updates only the continuation for $s\geq m+2$ loops.
Although each token traverses the shared stack once during decoding, recursive prefix memory propagation, current-loop depth KV, and U-MoLT's separate decoder preserve multi-loop prefix computation rather than collapsing to single-loop depth.
Even with an equal split $L_E=L_D=L/2$, U-MoLT reaches at least $3L/2$ blocks of causal path depth over earlier tokens, compared with at most $L$ in a non-looped $L$-layer Transformer.

Together, U-MoLT and loop KV close the inference efficiency gap in two steps.
First, even under fixed 2-loop execution, omitting unused decoder traversals lets equal-split U-MoLT match a non-looped $L$-layer Transformer in full-prompt prefill block depth ($2L_E=L$) and quadratic prefill attention, while keeping decoding traversals and persistent KV storage at $1.5\times$ instead of $2\times$.
Second, switching to MoL or MoL-Hybrid decoding lowers both an $L$-layer MoLT-KV stack and any U-MoLT partition with $L_E+L_D=L$ to $L$ block traversals per decoding step and a single persistent prefix KV history for the shared stack, matching a non-looped $L$-layer Transformer in total persistent KV storage.
Under MoL-Hybrid-2, equal-split U-MoLT therefore matches a non-looped backbone across all four dominant inference costs while retaining parallel prefill.
Its remaining decoding overhead consists of $L_E$ extra memory KV projections and small depth attention, which scales as an $O(L_E/T)$ fraction of sequence attention under full history.
During 2-loop prefill, it also adds the single-position decoder readout at $P-1$, prompt decoder KV projections, and extra memory KV projections.
\Cref{sec:analysis} provides the formal proofs, loop-forcing alignment analysis, and complete traversal, KV-storage, and computational cost breakdowns for the all-layer loop KV architectures, and \cref{sec:execution_results} evaluates these schedules empirically.
\section{Experiments}
\label{sec:experiments}
\suppressfloats[t]
The all-layer MoLA architecture in \cref{sec:method} provides the reference for our ablations.
We begin with S-size ablations of MoLT components and training choices.
The scaling experiments evaluate model quality in larger dense and MoE models.
The inference experiments compare teacher-forced prediction quality and execution cost across schedules.
Finally, we compare the continued pretraining performance of Qwen3.5-0.8B-Base with and without our MoLT adaptation.

\subsection{Experimental Setup}
\label{sec:experiment_setup}
\para{Models and training budgets}
Following the GPT-S scale convention~\citep{radford2019gpt2,brown2020gpt3}, the S-size ($\sim$113M) studies use an OLMo3-style pre-norm Transformer backbone of width 768 with grouped-query attention~\citep{ainslie2023gqa}, QK normalization, and SwiGLU, organized as either a plain 12-layer model or a 6-layer loop stack followed by a 6-layer decoder.
Their training pipeline uses GPT-2 tokenization, a 1,024-token context, and 3B training tokens from the FineWeb dataset~\citep{penedo2024fineweb}.

The scaling experiments cover 700M and 1.5B dense models and 3B MoE models with 660M active parameters, using 400B training tokens.
All these experiments use the OLMo3 engineering pipeline~\citep{olmoteam2025olmo3}, the \code{OLMo-mix-0625-official} mixture, Dolma2 tokenization, and 4,096-token contexts.
The Qwen adaptation starts from Qwen3.5-0.8B-Base~\citep{qwen2026qwen35base} and continues pretraining each compared architecture on 100B input tokens using the same Qwen tokenizer and 4,096-token context.

\para{Evaluation metrics}
We evaluate downstream accuracy on 10 benchmarks: PIQA~\citep{bisk2020piqa}, HellaSwag~\citep{zellers2019hellaswag}, WinoGrande~\citep{sakaguchi2020winogrande}, OpenBookQA~\citep{mihaylov2018openbookqa}, BoolQ~\citep{clark2019boolq}, SciQ~\citep{welbl2017sciq}, ARC-Easy (ARC-E) and ARC-Challenge (ARC-C)~\citep{clark2018arc}, COPA~\citep{roemmele2011copa}, and MMLU~\citep{hendrycks2021mmlu}.
We report final-checkpoint accuracy for each task and the arithmetic mean across tasks.

\begin{table}[!t]
  \centering
  \caption{\textbf{Architecture component ablations on S-size models.}
  Here \emph{cur.} abbreviates current and \emph{X} abbreviates cross.
  Depth source $\times$ disables depth KV, while \emph{cur. layer} reads only own KV.
  \emph{X layer} adds earlier current-loop local KV, and \emph{X loop} further adds all preceding-loop local KV.
  Both $\Delta$ Params and $\Delta$ FLOPs are relative to the same-loop baseline, namely method (1) for 1 loop and method (2) for 2 loops.
  Bold marks the lowest validation CE in each block at update 5,722.
  Step reports median maximum-rank update time over the final 500 updates.}
  \label{tab:components}
  \small
  \setlength{\tabcolsep}{2.5pt}
  \renewcommand{\arraystretch}{0.99}
  \begin{tabular*}{\linewidth}{@{\extracolsep{\fill}}lccccccccc@{}}
 \toprule
 Method & Num. & Feedback & \multicolumn{2}{c}{MoLA kernel} & Decoder & $\Delta$ Params & $\Delta$ FLOPs & CE & Step \\
 ID & Loops & form & Depth source & Placement & KV & (M) & (\%) & loss $\downarrow$ & (ms) $\downarrow$ \\
 \midrule
 \multicolumn{10}{@{}l}{\textbf{OLMo3 Baseline} (12 layers)} \\
 (1) & 1 & $\times$ & $\times$ & $\times$ & $\times$ & $0.000$ & $0.00$ & 3.318 & 494 \\ %
 (2) & 2 & Residual & $\times$ & $\times$ & $\times$ & $0.000$ & $0.00$ & \textbf{3.299} & 942 \\ %
 \midrule
 \multicolumn{10}{@{}l}{\textbf{MoLT-R} (12 layers)} \\
 (3) & 2 & Residual & X layer & dense & $\times$ & $0.000$ & $+0.08$ & 3.293 & 984 \\ %
 (4) & 2 & Residual & X layer & sparse & $\times$ & $0.000$ & $+0.03$ & 3.295 & 965 \\ %
 (5) & 2 & Residual & X loop & dense & $\times$ & $0.000$ & $+0.16$ & \textbf{3.284} & 1,029 \\ %
 (6) & 2 & Residual & X loop & sparse & $\times$ & $0.000$ & $+0.08$ & 3.286 & 984 \\ %
 \midrule
 \multicolumn{10}{@{}l}{\textbf{MoLT-KV} (12 layers)} \\
 (7) & 2 & KV & $\times$ & dense & $\times$ & $+3.540$ & $0.00$ & 3.322 & 968 \\ %
 (8) & 2 & KV & $\times$ & sparse & $\times$ & $+1.770$ & $0.00$ & 3.310 & 954 \\ %
 (9) & 2 & KV & cur. layer & dense & $\times$ & $+3.540$ & $+1.36$ & 3.308 & 998 \\ %
 (10) & 2 & KV & cur. layer & sparse & $\times$ & $+1.770$ & $+0.68$ & 3.296 & 973 \\ %
 (11) & 2 & KV & X layer & dense & $\times$ & $+3.540$ & $+1.43$ & 3.302 & 1,051 \\ %
 (12) & 2 & KV & X layer & sparse & $\times$ & $+1.770$ & $+0.71$ & 3.291 & 997 \\ %
 (13) & 2 & KV & X loop & dense & $\times$ & $+3.540$ & $+1.52$ & 3.297 & 1,101 \\ %
 (14) & 2 & KV & X loop & sparse & $\times$ & $+1.770$ & $+0.75$ & \textbf{3.286} & 1,018 \\ %
 \midrule
 \multicolumn{10}{@{}l}{\textbf{U-MoLT} (6+6 layers)} \\
 (15) & 2 & KV & cur. layer & dense & self & $+1.771$ & $+0.68$ & 3.300 & 969 \\ %
 (16) & 2 & KV & cur. layer & dense & shared & $+0.296$ & $-0.44$ & 3.297 & 951 \\ %
 (17) & 2 & KV & cur. layer & dense & re-proj. & $+1.771$ & $+0.68$ & 3.282 & 968 \\ %
 (18) & 2 & KV & cur. layer & sparse & self & $+0.886$ & $+0.34$ & 3.298 & 956 \\ %
 (19) & 2 & KV & cur. layer & sparse & shared & $-0.589$ & $-0.78$ & 3.293 & 935 \\ %
 (20) & 2 & KV & cur. layer & sparse & re-proj. & $+0.886$ & $+0.34$ & 3.282 & 964 \\ %
 (21) & 2 & KV & X layer & dense & self & $+1.771$ & $+0.70$ & 3.293 & 986 \\ %
 (22) & 2 & KV & X layer & dense & shared & $+0.296$ & $-0.42$ & 3.291 & 959 \\ %
 (23) & 2 & KV & X layer & dense & re-proj. & $+1.771$ & $+0.70$ & 3.279 & 982 \\ %
 (24) & 2 & KV & X layer & sparse & self & $+0.886$ & $+0.35$ & 3.296 & 969 \\ %
 (25) & 2 & KV & X layer & sparse & shared & $-0.589$ & $-0.77$ & 3.289 & 940 \\ %
 (26) & 2 & KV & X layer & sparse & re-proj. & $+0.886$ & $+0.35$ & 3.279 & 965 \\ %
 (27) & 2 & KV & X loop & dense & self & $+1.771$ & $+0.72$ & 3.294 & 995 \\ %
 (28) & 2 & KV & X loop & dense & shared & $+0.296$ & $-0.40$ & 3.291 & 976 \\ %
 (29) & 2 & KV & X loop & dense & re-proj. & $+1.771$ & $+0.72$ & \textbf{3.273} & 1,001 \\ %
 (30) & 2 & KV & X loop & sparse & self & $+0.886$ & $+0.36$ & 3.294 & 967 \\ %
 (31) & 2 & KV & X loop & sparse & shared & $-0.589$ & $-0.76$ & 3.285 & 947 \\ %
 (32) & 2 & KV & X loop & sparse & re-proj. & $+0.886$ & $+0.36$ & 3.276 & 968 \\ %
 \bottomrule
 \end{tabular*}
  \end{table}

For language modeling, we report cross-entropy (CE) and perplexity (PPL), where lower values are better.
The scaling experiments evaluate PPL on C4~\citep{raffel2020t5}, ICE~\citep{greenbaum1996ice}, M2D2-S2ORC~\citep{reid2022m2d2,lo2020s2orc}, the Pile~\citep{gao2020pile}, WikiText-103~\citep{merity2016wikitext}, and 5 Dolma domains~\citep{soldaini2024dolma}: Books, Common Crawl, peS2o, Reddit, and Stack.
We report per-domain PPL and the arithmetic mean across these 10 domains.
For Qwen continued pretraining, we evaluate CE on fixed validation blocks across 11 domains, adding Dolma Wiki to the 10 domains above, and report their mean.

\subsection{Component Ablations}
\label{sec:components}
\label{sec:plain}
\suppressfloats[t]
We study how feedback, depth sources, and decoder design affect model quality under a common training protocol.
\Cref{tab:components} compares 32 S-size configurations in four blocks: OLMo3 Baseline, MoLT-R, MoLT-KV, and U-MoLT.
All models are trained on 3B input tokens in BF16 with Muon, using seed 11 and the same parameter initialization.
Two-loop models use equal loop-loss weights, i.e., 1:1.
We report final-loop CE, update time, and parameter and FLOP differences relative to the baseline with the same loop count: method (1) for 1 loop and method (2) for 2 loops.

The first three blocks in \cref{tab:components} use complete 12-layer stacks, while U-MoLT splits the stack into a 6-layer encoder and a 6-layer decoder.
The depth-source choices progressively expand same-token local KV access: \emph{cur. layer} reads the current layer, \emph{X layer} adds cross-layer sources in the current loop, and \emph{X loop} further adds cross-loop sources.
For MoLT-KV and U-MoLT, \emph{X loop} is an extended ablation beyond the current-loop default in \cref{sec:mola}.
Conversely, for MoLT-R, \emph{X layer} ablates the preceding-loop local KV reads included in \cref{sec:encoder,sec:mola}.
Modern LLMs~\citep{team2026kimik3} often use hybrid architectures that mix full attention with linear or sparse attention layers.
To test partial placement, we compare \emph{dense} placement, which applies MoLA (or strict-prefix KV feedback in no-depth controls) at every layer, with \emph{sparse} placement, which applies it at alternating layers while retaining ordinary causal self-attention at the remaining layers.
For U-MoLT, \emph{self} constructs decoder KV from each layer's own hidden states, while \emph{shared} and \emph{re-proj.} read the encoder output through shared and independent per-layer KV projections, respectively, with \emph{re-proj.} and current-loop \emph{X layer} corresponding to the \cref{sec:decoder} U-MoLT design.

\para{Looping and depth reuse improve prediction}
Residual looping improves prediction with unchanged parameter count.
Moving from the single-loop method (1) to the 2-loop method (2) lowers CE from 3.318 to 3.299, a gain of 0.019 at the same input-token budget.
Both models have 112.972M parameters.
The 2-loop recipe doubles logical training FLOPs and increases update time from 494 to 942 ms.

Direct depth access further strengthens residual feedback.
With dense placement, method (3) adds current-loop cross-layer sources to method (2) and lowers CE by 0.006.
Method (5) further adds preceding-loop sources and improves on method (2) by 0.015.
The preceding-loop gain also holds with sparse placement, where method (6) improves on method (2) by 0.013.
Adding depth KV therefore lowers the best MoLT-R CE to 3.284 without adding parameters.

Expanding depth KV sources also improves MoLT-KV under both placements.
Without depth KV, methods (7) and (8) reduce feedback layers to strict-prefix cross-attention that cannot read KV at the current position.
Both perform worse than the 2-loop baseline (2), and dense method (7) even performs worse than the single-loop baseline (1).
CE decreases steadily from no depth to current-layer, cross-layer, and cross-loop access.
Adding own KV provides a direct same-position attention path alongside strict-prefix memory KV, complementing the residual and FFN paths.
Reading earlier layers and the preceding loop brings further gains without adding parameters.
In \cref{tab:components}, no-depth methods (7) and (8) show $0.00\%$ $\Delta$FLOPs because loop~2 projects only memory KV in place of local KV, whereas adding own KV computes both projections in loop~2 ($+1.36\%$ dense, $+0.68\%$ sparse) and expanding further to \emph{X layer} or \emph{X loop} reuses those local KV with negligible extra attention FLOPs.

\para{Placement depends on the feedback form}
Dense placement gives the best MoLT-R quality, while sparse placement gives MoLT-KV both lower CE and faster updates in all 4 source conditions.
For MoLT-KV, sparse placement reduces CE by 0.011--0.012 relative to dense placement and saves 1.770M parameters.
Comparing the two feedback forms, MoLT-R outperforms MoLT-KV with dense cross-loop sources, while MoLT-KV outperforms MoLT-R with sparse cross-layer sources.
These results make feedback form and placement a joint design choice, with sparse KV feedback retaining ordinary self-attention between feedback layers.

\para{Independent decoder KV projections improve U-MoLT}
The U-shaped design allows decoder KV sources and projections to be chosen separately from encoder feedback.
With ordinary decoder self-attention, moving from full-stack MoLT-KV to a 6+6 split improves all 3 dense source conditions but raises CE in their sparse counterparts.
This split changes the feedback boundary from layer 11 to layer 5 and assigns the remaining 6 layers to prediction.
We therefore compare decoder KV sources within each fixed U-MoLT encoder configuration.

\begin{table}[!t]
  \centering
  \caption{\textbf{Effect of loop-loss weights on per-loop validation CE.}
  All runs use S-size U-MoLT with sparse MoLA placement, full-history current-loop local KV, and 3B training tokens.
  Loss weight gives the ratio of loop 1 to loop 2 loss weights.
  Bold marks the lowest CE for each loop.}
  \label{tab:loop_loss_weights}
  \small
  \setlength{\tabcolsep}{12pt}
  \begin{tabular}{lcccccc}
  \toprule
  \multirow{2}{*}{Loss weight} & \multicolumn{2}{c}{1:1} & \multicolumn{2}{c}{1:2} & \multicolumn{2}{c}{1:4} \\
  \cmidrule(lr){2-3}\cmidrule(lr){4-5}\cmidrule(lr){6-7}
  & Loop 1 & Loop 2 & Loop 1 & Loop 2 & Loop 1 & Loop 2 \\
  \midrule
  CE loss $\downarrow$ & \textbf{3.306} & 3.279 & 3.315 & 3.276 & 3.322 & \textbf{3.272} \\
  \bottomrule
  \end{tabular}
  \end{table}

Independent decoder KV projections give the most consistent gain in the component study.
All 6 matched triplets of decoder KV configurations follow the same CE ordering: \emph{re-proj.} is best, followed by \emph{shared} and then \emph{self}.
Relative to \emph{self}, \emph{re-proj.} lowers CE by 0.014--0.021 with the same parameter count and less than 1\% change in update time.
Sharing the projections saves 1.475M parameters but increases CE by 0.009--0.018 relative to \emph{re-proj.}
Because \emph{shared} computes one decoder KV projection instead of 6 per loop, it also yields negative $\Delta$FLOPs relative to the 2-loop baseline.
The consistent ordering supports separating encoder memory construction from decoder updates, with each decoder layer projecting the normalized encoder output independently.
We discuss this decoupling further in \cref{sec:discussion}.

\para{Depth sources and decoder KV configuration work together}
Within U-MoLT, current-loop cross-layer access improves all 6 combinations of placement and decoder KV configuration, and preceding-loop access further improves both \emph{re-proj.} configurations.
Under dense placement, adding preceding-loop sources to \emph{re-proj.} lowers CE by 0.006 from method (23) to method (29).
For \emph{self}, the same change from method (21) to method (27) does not improve CE.
Across the 9 matched dense and sparse U-MoLT pairs, dense placement gives lower CE in 2 pairs, sparse placement gives lower CE in 4, and 3 tie, showing no consistent advantage for dense placement.

Dense cross-loop \emph{re-proj.} in method (29) achieves the lowest CE in the full matrix, reaching 3.273 at 1,001 ms/update.
It adds 1.771M parameters and 0.72\% logical training FLOPs relative to the 2-loop baseline.
Sparse cross-loop \emph{re-proj.} in method (32) offers a faster alternative at 3.276 CE and 968 ms/update.
Dense cross-layer \emph{re-proj.} in method (23) reaches 3.279 at 982 ms/update.
These choices expose the quality and update-cost tradeoff of depth information reuse.
Because the cross-loop variants read preceding-loop local KV at the same token, only the current-loop variants support single-traversal MoL decoding.

\para{Greater final-loop loss weight improves final-loop prediction}
We compare first-to-second-loop loss weights of 1:1, 1:2, and 1:4 in \cref{tab:loop_loss_weights}.
The comparison uses the same architecture and 3B-token training recipe as method (26) in \cref{tab:components}.

Increasing the relative weight on the second-loop loss consistently lowers second-loop CE but raises first-loop CE across the three settings.
Changing the weights from 1:1 to 1:4 lowers second-loop CE from 3.279 to 3.272, while first-loop CE rises from 3.306 to 3.322.
The component analysis in \cref{tab:components} uses 1:1 weights to compare architectures under a common learning objective.
We next evaluate model quality at larger dense and MoE scales.

\FloatBarrier

\FloatBarrier
\begingroup
\interlinepenalty=10000
\subsection{Scaling to Dense and MoE Models}
\label{sec:last10}
\label{sec:large}
\suppressfloats[t]
We evaluate whether MoLT's quality gains extend to larger dense and MoE models at a common budget of 400B input tokens.
The comparison covers 700M and 1.5B dense models and 3B MoE models with 660M active parameters.
\Cref{tab:last10_accuracy,tab:last10_ppl} report final-checkpoint accuracy on 10 tasks and perplexity on 10 text domains, respectively, with averaged scores across tasks or domains.
For MMLU, we report accuracy under the MMLU Var protocol.
All results are evaluated at the final checkpoint using final-loop predictions for the looped models, while single-traversal MoL decoding is evaluated separately in \cref{sec:execution_results}.

\begin{table}[H]
\centering
\caption{\textbf{Final-checkpoint downstream accuracy.}
Values are percentages, where higher is better.
Avg.\ is the mean across the 10 tasks.
MMLU accuracy is evaluated under the MMLU Var protocol.
Bold marks the highest score in each model block.}
\label{tab:last10_accuracy}
\begingroup
\small
\setlength{\tabcolsep}{2.5pt}
\renewcommand{\arraystretch}{0.99}
\begin{tabular*}{\textwidth}{@{\extracolsep{\fill}}lccccccccccc@{}}
\toprule
\multirow{2}{*}{Method} & \multirow{2}{*}{PIQA} & Hella & Wino & OpenBook & \multirow{2}{*}{BoolQ} & \multirow{2}{*}{SciQ} & \multirow{2}{*}{ARC-E} & \multirow{2}{*}{ARC-C} & \multirow{2}{*}{COPA} & \multirow{2}{*}{MMLU} & \multirow{2}{*}{Avg. $\uparrow$} \\
 & & Swag & Grande & QA & & & & & & & \\
\midrule
\multicolumn{12}{@{}l}{\textbf{700M dense models}} \\
OLMo3 & 70.7 & 53.1 & 55.8 & 36.2 & 56.0 & 89.3 & 66.4 & 35.4 & 69.0 & 33.9 & 56.6 \\
MoLT-R & \textbf{72.5} & \textbf{55.8} & \textbf{58.0} & 36.2 & \textbf{62.0} & \textbf{92.5} & \textbf{69.6} & \textbf{38.1} & 73.0 & 34.3 & \textbf{59.2} \\
U-MoLT & 72.4 & 55.0 & 56.0 & \textbf{37.2} & 59.4 & 90.2 & 69.1 & 35.5 & \textbf{81.0} & \textbf{35.4} & 59.1 \\
\midrule
\multicolumn{12}{@{}l}{\textbf{1.5B dense models}} \\
OLMo3 & 73.9 & 59.9 & 58.4 & 37.4 & 63.7 & 91.8 & \textbf{74.2} & 40.4 & 76.0 & 36.6 & 61.2 \\
U-MoLT & \textbf{74.8} & \textbf{61.8} & \textbf{59.0} & \textbf{39.0} & \textbf{65.7} & \textbf{93.0} & \textbf{74.2} & \textbf{42.1} & \textbf{79.0} & \textbf{37.0} & \textbf{62.6} \\
\midrule
\multicolumn{12}{@{}l}{\textbf{3B MoE models, 660M active parameters}} \\
OLMo3 & 72.9 & 59.3 & 59.5 & 35.4 & 63.1 & 89.8 & 70.1 & 36.7 & 77.0 & 36.7 & 60.1 \\
U-MoLT & \textbf{73.2} & \textbf{62.0} & \textbf{59.9} & \textbf{37.6} & \textbf{64.4} & \textbf{90.5} & \textbf{73.9} & \textbf{41.8} & \textbf{80.0} & \textbf{38.2} & \textbf{62.2} \\
\bottomrule
\end{tabular*}
\endgroup
\end{table}

\begin{table}[H]
\centering
\caption{\textbf{Final-checkpoint validation perplexity across 10 domains.}
Lower is better, and bold marks the lowest perplexity in each model block.
Books, CC, peS2o, Reddit, and Stack are 5 Dolma domains, where CC denotes Common Crawl.
Avg.\ is the mean across the 10 domains.}
\label{tab:last10_ppl}
\begingroup
\small
\setlength{\tabcolsep}{2.5pt}
\renewcommand{\arraystretch}{0.99}
\begin{tabular*}{\textwidth}{@{\extracolsep{\fill}}lccccccccccc@{}}
\toprule
\multirow{2}{*}{Method} & \multirow{2}{*}{C4} & \multirow{2}{*}{ICE} & M2D2- & \multirow{2}{*}{Pile} & WikiText- & \multirow{2}{*}{Books} & \multirow{2}{*}{CC} & \multirow{2}{*}{peS2o} & \multirow{2}{*}{Reddit} & \multirow{2}{*}{Stack} & \multirow{2}{*}{Avg. $\downarrow$} \\
 & & & S2ORC & & 103 & & & & & & \\
\midrule
\multicolumn{12}{@{}l}{\textbf{700M dense models}} \\
OLMo3 & 18.68 & 18.59 & 21.25 & 9.43 & 12.36 & 16.50 & 21.53 & 8.78 & 25.51 & 3.75 & 15.64 \\
MoLT-R & \textbf{17.82} & 17.59 & 19.77 & \textbf{9.04} & 11.70 & \textbf{15.63} & \textbf{20.53} & 8.47 & \textbf{24.26} & 3.63 & \textbf{14.84} \\
U-MoLT & 17.89 & \textbf{17.44} & \textbf{19.75} & \textbf{9.04} & \textbf{11.56} & 15.74 & 20.62 & \textbf{8.44} & 24.48 & \textbf{3.62} & 14.86 \\
\midrule
\multicolumn{12}{@{}l}{\textbf{1.5B dense models}} \\
OLMo3 & 16.53 & 16.72 & 18.75 & 8.49 & 10.62 & 14.23 & 19.05 & 7.94 & 22.77 & 3.45 & 13.86 \\
U-MoLT & \textbf{15.81} & \textbf{15.52} & \textbf{17.77} & \textbf{8.11} & \textbf{9.92} & \textbf{13.50} & \textbf{18.16} & \textbf{7.63} & \textbf{21.71} & \textbf{3.34} & \textbf{13.15} \\
\midrule
\multicolumn{12}{@{}l}{\textbf{3B MoE models, 660M active parameters}} \\
OLMo3 & 16.61 & 17.67 & 18.42 & 8.48 & 10.91 & 14.32 & 19.11 & 7.98 & 22.93 & 3.44 & 13.99 \\
U-MoLT & \textbf{15.76} & \textbf{16.47} & \textbf{17.59} & \textbf{8.08} & \textbf{10.05} & \textbf{13.53} & \textbf{18.16} & \textbf{7.64} & \textbf{21.82} & \textbf{3.31} & \textbf{13.24} \\
\bottomrule
\end{tabular*}
\endgroup
\end{table}

\para{Quality gains extend across model scales}
U-MoLT improves mean accuracy and lowers perplexity at all three scales.
Relative to OLMo3, average accuracy increases by 2.5, 1.4, and 2.1 percentage points at 700M dense, 1.5B dense, and 3B MoE, respectively.
Perplexity improves in each of the 10 domains at each scale, including a reduction in mean PPL from 13.99 to 13.24 for the MoE model.
The downstream gains are similarly broad: U-MoLT improves 9 tasks and ties ARC-E at 1.5B, and improves all 10 tasks on the 3B MoE model.

\para{Both feedback forms improve the 700M baseline}
MoLT-R and U-MoLT both improve the 700M model, reaching mean accuracies of 59.2\% and 59.1\% against OLMo3's 56.6\%.
Their mean perplexities fall to 14.84 and 14.86 from 15.64, with improvements in all 10 domains.
MoLT-R achieves the best aggregate scores in this block, while U-MoLT leads on OpenBookQA, COPA, and MMLU.
This contrast reflects their complementary roles.
MoLT-R prioritizes quality under fixed-loop execution, whereas U-MoLT delivers nearly matching quality while supporting the MoL decoding schedule analyzed in \cref{sec:analysis}.
\par
\endgroup

\para{Faster convergence with fewer training tokens}
On a paired 700M/400B run using a separate split of \code{OLMo-mix-0625-official} with periodic 11-domain evaluation (adding Dolma Wiki), U-MoLT matches OLMo3's 400B-token mean validation CE at 232.8B tokens ($1.7\times$ token efficiency) and achieves a $1.5\times$ gain on smoothed final-loop training CE.
At 400B tokens, U-MoLT lowers CE across all 11 domains, reducing the equal-domain mean from 2.598 to 2.553 (\cref{sec:large_details}, \cref{fig:large_loss_speedup,fig:large_val_ce}).

\FloatBarrier
\subsection{MoL Decoding: Quality and Efficiency}
\label{sec:execution_results}
\label{sec:readout_results}
\suppressfloats[t]
We compare the prediction quality and latency of fixed 2-loop, Mixture-of-Loops (MoL), MoL-Hybrid-1, and MoL-Hybrid-2 decoding (\cref{sec:schedules}).
Fixed 2-loop decoding runs two traversals per step, MoL decoding runs one traversal per token from empty caches, and MoL-Hybrid-1 and MoL-Hybrid-2 pair 1- and 2-loop parallel prefill with the MoL decoding update.
This comparison separates the benefits of parallel prefill and single-traversal decoding.

Each model is evaluated under all four schedules using the same weights and continuation targets.
The main comparison uses a separate pair of 3B-token S-size MoLT-KV and U-MoLT checkpoints trained with detached feedback gradients, distinct from the open-gradient ablations in \cref{tab:components}.
MoLT-KV applies memory feedback at alternating layers (0, 2, \ldots, 10) while retaining current-loop depth access across all 12 layers, whereas U-MoLT uses dense memory feedback across its 6-layer encoder.
Although \cref{sec:analysis} analyzes all-layer MoLA stacks, the same stabilization bounds extend to sparse feedback placement (\cref{sec:prefix_stabilization}).
Non-feedback layers read causal current-loop sequence KV at $j\leq t$, together with current-loop depth KV at $t$ when enabled, without directly reading preceding-loop KV.
Their states and current-loop sequence KV caches therefore stabilize jointly with $z^r_t$ at $r\geq t+2$.
For fixed-prompt continuations, both MoL-Hybrid decoding and the fixed-prompt continuation unroll (\cref{cor:hybrid_stabilization}) also freeze and reuse each non-feedback layer's prompt current-loop sequence KV from the final prefill loop.
This preserves the $s\geq m+2$ bound.
Under single-traversal decoding, each non-feedback layer retains one current-loop sequence KV cache in place of a memory KV cache, preserving the same total persistent KV cache payload (\cref{sec:cost}) with fewer memory projections.
We score 8 fixed windows, each with a 128-token prompt used only as context.
CE is computed on the following 896 teacher-forced continuation targets per window, totaling 7,168 scored targets per schedule.
The first target is predicted from the final prefill logits, and the remaining 895 targets are predicted during decoding.
\Cref{tab:decoding_quality_latency} reports CE, PPL, and their differences from fixed 2-loop decoding alongside total execution time on one H20 GPU at batch size 1.

\begin{table}[htbp]
\centering
\caption{\textbf{Quality and total latency on matched continuation text.}
Each model is evaluated from the same checkpoint across all four decoding schedules on 8 windows totaling 7,168 targets.
MoL-Hybrid-1 and MoL-Hybrid-2 use 1- and 2-loop parallel prefill followed by MoL decoding.
Both $\Delta$CE and $\Delta$PPL are differences from fixed 2-loop decoding computed from unrounded values.
Per-window total time covers the 128-token prefill and 895 decoding steps, averaged across the 8 windows (5 runs per window) on one H20 GPU at batch size 1.}
\label{tab:decoding_quality_latency}
\small
\setlength{\tabcolsep}{4pt}
\begin{tabular*}{\textwidth}{@{\extracolsep{\fill}}llrrrrr@{}}
\toprule
Method & Decoding schedule & CE $\downarrow$ & PPL $\downarrow$ & $\Delta$CE & $\Delta$PPL & Total (s) \\
\midrule
MoLT-KV & Fixed 2-loop & 3.320 & 27.67 & $0.0000$ & $0.000$ & 31.510 \\
MoLT-KV & Mixture-of-Loops & 3.321 & 27.68 & $+0.0002$ & $+0.006$ & 19.109 \\
MoLT-KV & MoL-Hybrid-1 & 3.322 & 27.71 & $+0.0014$ & $+0.039$ & 16.786 \\
MoLT-KV & MoL-Hybrid-2 & 3.321 & 27.68 & $+0.0002$ & $+0.005$ & 16.852 \\
\midrule
U-MoLT & Fixed 2-loop & 3.280 & 26.58 & $0.0000$ & $0.000$ & 23.430 \\
U-MoLT & Mixture-of-Loops & 3.280 & 26.58 & $+0.0002$ & $+0.006$ & 18.143 \\
U-MoLT & MoL-Hybrid-1 & 3.280 & 26.57 & $-0.0002$ & $-0.004$ & 16.053 \\
U-MoLT & MoL-Hybrid-2 & 3.280 & 26.56 & $-0.0005$ & $-0.012$ & 16.873 \\
\bottomrule
\end{tabular*}
\end{table}

\para{MoL-Hybrid decoding reduces latency with small perplexity changes}
MoL-Hybrid decoding reduces total latency, with $\Delta$CE between $-0.0005$ and $+0.0014$ and $\Delta$PPL between $-0.012$ and $+0.039$.
For MoLT-KV, MoL-Hybrid-1 and MoL-Hybrid-2 reduce total time by 46.7\% and 46.5\% relative to fixed 2-loop decoding, with $\Delta$PPL of $+0.039$ and $+0.005$, respectively.
For U-MoLT, the time reductions are 31.5\% and 28.0\%, while PPL decreases by 0.004 and 0.012.
MoL-Hybrid-1 gives the lowest latency for both checkpoints, while MoL-Hybrid-2 gives the lower PPL of the two variants.
MoL decoding takes longer overall than either MoL-Hybrid variant because it processes the prompt sequentially.

\para{Fewer traversals reduce decoding time}
A separate timing study isolates prefill and decoding costs for 128-token and 4,096-token prompts (\cref{tab:decoding_phase_latency}).
Each request runs a full prompt prefill followed by 64 single-token decoding steps, producing 65 continuation predictions.
The 128-token entries and both 4,096-token Mixture-of-Loops entries are measured on one H20 GPU in BF16 at batch size 1.
The 9 fixed-loop and MoL-Hybrid entries at 4,096 tokens report the summary timings plotted in \cref{fig:results_overview}(b).

\begin{table}[!t]
\centering
\caption{\textbf{Prefill and decoding latency for short and long prompts.}
Times are reported in seconds for prompt prefill followed by 64 single-token decoding steps.
The 128-token entries and both 4,096-token Mixture-of-Loops entries run on one H20 GPU in BF16 at batch size 1.
The 9 fixed-loop and MoL-Hybrid entries at 4,096 tokens match the summary timings in \cref{fig:results_overview}(b).
MoL-Hybrid-1 and MoL-Hybrid-2 use 1- and 2-loop parallel prefill followed by MoL decoding.
}
\label{tab:decoding_phase_latency}
\small
\setlength{\tabcolsep}{3pt}
\begin{tabular*}{\textwidth}{@{\extracolsep{\fill}}llrrrrrr@{}}
\toprule
\multirow{2}{*}[-0.5\dimexpr\aboverulesep+\belowrulesep+\cmidrulewidth\relax]{Method} & \multirow{2}{*}[-0.5\dimexpr\aboverulesep+\belowrulesep+\cmidrulewidth\relax]{Schedule} & \multicolumn{3}{c}{Prompt: 128 tokens} & \multicolumn{3}{c}{Prompt: 4,096 tokens} \\
\cmidrule(lr){3-5}\cmidrule(lr){6-8}
 & & Prefill & Decode & Total & Prefill & Decode & Total \\
\midrule
OLMo3 & Fixed 1-loop & 0.0129 & 0.9776 & 0.9905 & 0.0181 & 0.9825 & 1.0006 \\
Loop OLMo3 & Fixed 2-loop & 0.0245 & 1.9617 & 1.9862 & 0.0347 & 1.9396 & 1.9743 \\
\midrule
MoLT-R & Fixed 2-loop & 0.0256 & 1.8707 & 1.8963 & 0.0356 & 1.9412 & 1.9768 \\
\midrule
MoLT-KV & Fixed 2-loop & 0.0297 & 1.8290 & 1.8587 & 0.0362 & 1.9439 & 1.9801 \\
MoLT-KV & Mixture-of-Loops & 2.2539 & 0.9370 & 3.1909 & 72.8810 & 0.9592 & 73.8402 \\
MoLT-KV & MoL-Hybrid-1 & 0.0161 & 0.9542 & 0.9703 & 0.0193 & 0.9702 & 0.9895 \\
MoLT-KV & MoL-Hybrid-2 & 0.0328 & 0.9650 & 0.9978 & 0.0378 & 0.9688 & 1.0066 \\
\midrule
U-MoLT & Fixed 2-loop & 0.0242 & 1.4291 & 1.4533 & 0.0245 & 1.4430 & 1.4675 \\
U-MoLT & Mixture-of-Loops & 1.4386 & 0.9509 & 2.3895 & 46.2151 & 0.9633 & 47.1784 \\
U-MoLT & MoL-Hybrid-1 & 0.0182 & 0.9674 & 0.9856 & 0.0183 & 0.9712 & 0.9895 \\
U-MoLT & MoL-Hybrid-2 & 0.0268 & 0.9675 & 0.9944 & 0.0265 & 0.9701 & 0.9966 \\
\bottomrule
\end{tabular*}
\end{table}

\begin{table}[!t]
  \centering
  \caption{\textbf{Qwen3.5-0.8B-Base adaptation after 100B tokens of continued pretraining.}
  Qwen3.5-CT retains the original architecture, while both U-MoLT adaptations use \emph{re-proj.} decoder attention.
  U-MoLT-CT1 builds decoder GDN memory from each layer's own input, whereas U-MoLT-CT2 builds it from the encoder output.
  All models are evaluated on fixed validation blocks, with bold marking the lowest CE per domain.
  CC denotes Common Crawl, and Avg.\ is the mean across the 11 domains.}
  \label{tab:qwen_ct}
  \label{tab:qwen_ct_domains}
  \begingroup
  \small
  \setlength{\tabcolsep}{2.5pt}
  \renewcommand{\arraystretch}{0.99}
  \begin{tabular*}{\textwidth}{@{\extracolsep{\fill}}lcccccccccccc@{}}
  \toprule
  \multirow{2}{*}{Method} & \multirow{2}{*}{C4} & \multirow{2}{*}{ICE} & M2D2- & \multirow{2}{*}{Pile} & WikiText- & \multirow{2}{*}{Books} & \multirow{2}{*}{CC} & \multirow{2}{*}{peS2o} & \multirow{2}{*}{Reddit} & \multirow{2}{*}{Stack} & \multirow{2}{*}{Wiki} & \multirow{2}{*}{Avg. $\downarrow$} \\
   & & & S2ORC & & 103 & & & & & & & \\
  \midrule
  Qwen3.5-CT & 2.759 & 2.967 & 2.995 & 2.057 & 2.401 & 2.710 & 2.939 & 2.039 & 3.170 & 1.042 & 2.376 & 2.496 \\
  U-MoLT-CT1 & \textbf{2.741} & \textbf{2.945} & \textbf{2.983} & \textbf{2.042} & \textbf{2.380} & \textbf{2.690} & \textbf{2.920} & \textbf{2.025} & \textbf{3.154} & \textbf{1.036} & \textbf{2.359} & \textbf{2.480} \\
  U-MoLT-CT2 & 2.745 & 2.946 & 2.985 & 2.045 & 2.383 & 2.694 & 2.923 & 2.028 & 3.157 & 1.040 & 2.363 & 2.483 \\
  \bottomrule
  \end{tabular*}
  \endgroup
  \end{table}

MoL and MoL-Hybrid decoding approximately halve MoLT-KV's decoding time by reducing block traversals from 24 to 12 per decoding step.
U-MoLT reduces them from 18 to 12: its fixed 2-loop schedule already omits the unused first-loop decoder and executes two 6-layer encoder traversals followed by one 6-layer decoder traversal.
This difference explains the smaller reduction in decoding time for U-MoLT.
Both architectures reach similar decoding times with a single traversal.
All models share the same backbone implementation, and MoLA layers use our fused MoLA kernel.
With equal block traversals per step, single-traversal MoLT decoding reaches latencies close to non-looped OLMo3, consistent with the small extra depth-attention and memory KV projection work in \cref{sec:cost}.

\para{Parallel prefill reduces total latency for long prompts}
MoL-Hybrid decoding keeps prompt processing fast as context length grows.
On the matched 128-token H20 measurements, MoL-Hybrid reduces total time relative to sequential MoL decoding from 3.1909 to 0.9703--0.9978\,s for MoLT-KV and from 2.3895 to 0.9856--0.9944\,s for U-MoLT.
At 4,096 tokens, sequential MoL decoding grows to 73.8402\,s for MoLT-KV and 47.1784\,s for U-MoLT on H20, whereas MoL-Hybrid totals remain near one second in the \cref{fig:results_overview}(b) summary at 0.9895--1.0066\,s and 0.9895--0.9966\,s.
Within that 4,096-token summary, MoL-Hybrid totals are 49.2--50.0\% lower than fixed 2-loop decoding for MoLT-KV and 32.1--32.6\% lower for U-MoLT.
Under MoL-Hybrid-2, U-MoLT also achieves shorter 4,096-token prefill than MoLT-KV by omitting decoder blocks at non-final prompt positions.

\begin{figure}[!t]
  \centering
  \includegraphics[width=\textwidth]{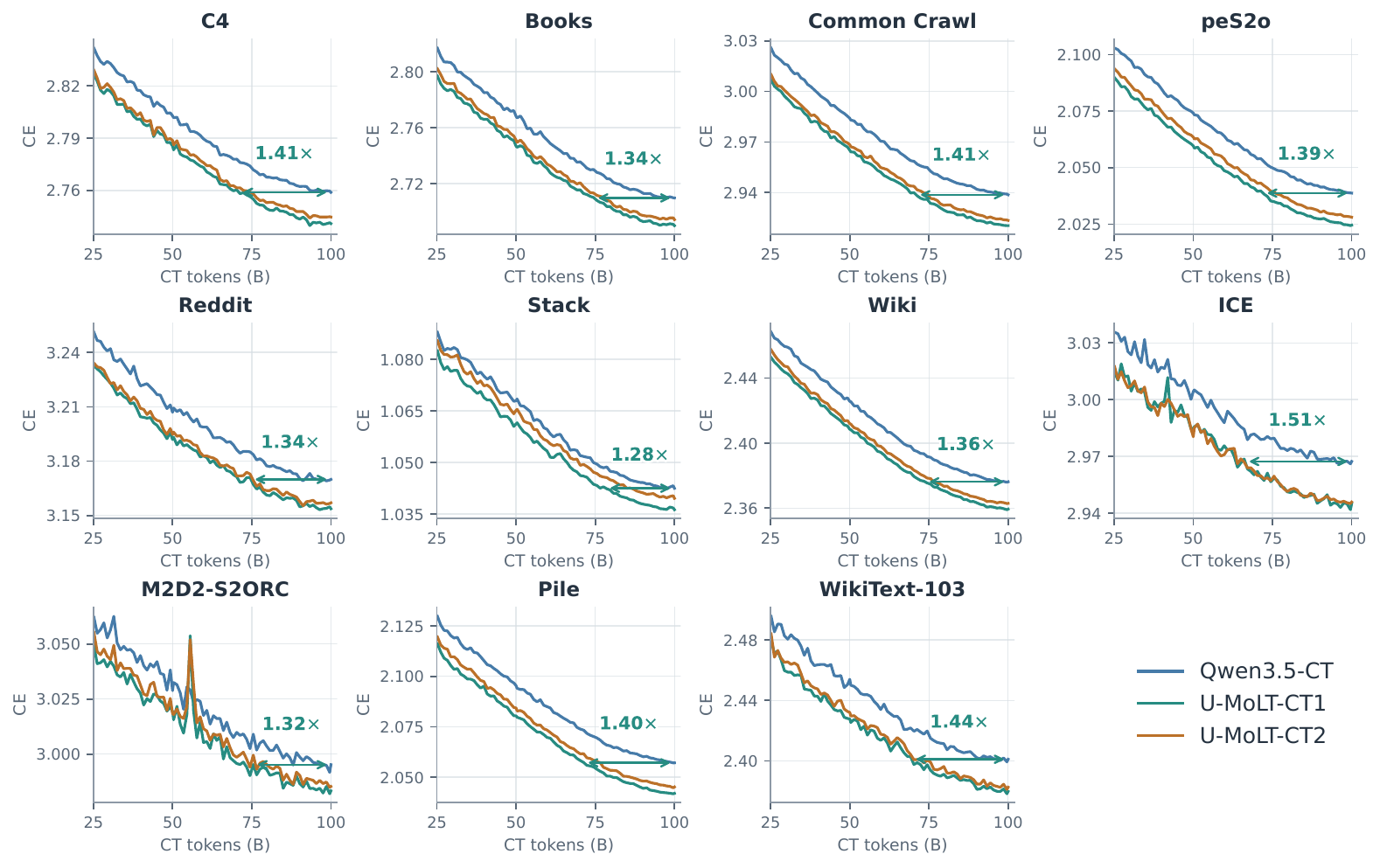}
  \caption{\textbf{Validation CE trajectories during 100B-token Qwen continued pretraining.}
  Each panel shows validation CE from 25B tokens onward.
  Teal arrows mark where U-MoLT-CT1 reaches Qwen3.5-CT's final CE via linear interpolation, and annotated factors report the corresponding input-token speedup.
  Both U-MoLT adaptations are evaluated from their second-loop predictions.}
  \label{fig:qwen_ct_100b}
  \end{figure}

\subsection{Adapting Qwen through Continued Pretraining}
\label{sec:qwen_ct}
We test whether the MoLT feedback interface improves an existing pretrained backbone through matched continued pretraining (CT).
Starting from Qwen3.5-0.8B-Base, we train Qwen3.5-CT, U-MoLT-CT1, and U-MoLT-CT2 on the same 100B-token stream with the same tokenizer and AdamW schedule.
Training uses a 1.05M-token global batch and a peak learning rate of $10^{-4}$, with a 1.05B-token warmup and cosine decay to $10^{-5}$.
Qwen3.5-CT retains the original architecture, while U-MoLT-CT1 and U-MoLT-CT2 apply two MoLT adaptations.
Evaluation uses fixed validation blocks from 11 domains and their mean CE, with final second-loop predictions for the adaptations.
\Cref{tab:qwen_ct} compares the three architectures after 100B tokens of continued pretraining.

U-MoLT-CT1 and U-MoLT-CT2 differ in the source used to construct decoder Gated DeltaNet (GDN)~\citep{yang2025gateddelta} memory.
Qwen3.5-0.8B-Base has 24 layers, comprising 18 GDN layers and 6 full-attention layers.
Both adaptations split this backbone into a 12-layer encoder and a 12-layer decoder, with 9 GDN layers and 3 full-attention layers in each half.
In the encoder, the 3 full-attention layers act as MoLA feedback layers reading strict-prefix memory KV from the preceding loop, while the 9 GDN layers remain recurrent sequence layers whose states reset between loops.
Both use the \emph{re-proj.} decoder-attention design in \cref{tab:components}, with independent decoder KV projections at each attention layer.
U-MoLT-CT1 constructs memory in the 9 decoder GDN layers from each layer's own input, following the \emph{self} source choice.
Because these decoder GDN layers retain dependence on earlier decoder states, U-MoLT-CT1 is not strictly cross-only like pure-attention U-MoLT.
U-MoLT-CT2 instead projects decoder GDN KV, decay gates, and update strengths from the normalized encoder output, while keeping queries and output gates dependent on the decoder state.
Because each Qwen3.5 GDN layer applies a length-4 causal convolution to its queries, U-MoLT-CT2 still retains short-range cross-position decoder dependence across its 9 decoder GDN layers (a 28-token trailing window) rather than exact single-position prefill omission.
The two adaptations share initialization, parameter count, encoder computation, and equal loop-loss weights.

\para{MoLT improves all 11 domains over the matched baseline}
Both adaptations achieve lower CE than Qwen3.5-CT in every validation domain after 100B tokens.
U-MoLT-CT1 reduces the equal-domain mean from 2.496 to 2.480, and U-MoLT-CT2 reaches 2.483 (\cref{tab:qwen_ct}).
These broad improvements show that loop feedback transfers to an existing pretrained backbone under the same input-token budget, using additional shared-weight loop computation.
Retaining decoder-state GDN writes gives the strongest adaptation.
U-MoLT-CT1 outperforms U-MoLT-CT2 in all 11 domains, with a mean CE advantage of 0.003.

\para{U-MoLT-CT1 reaches the baseline endpoint with fewer input tokens}
U-MoLT-CT1 matches Qwen3.5-CT's final domain-level CE after 66.3--78.3B input tokens along the recorded training trajectories (\cref{fig:qwen_ct_100b}).
Across the 11 domains, this yields a $1.28$--$1.51\times$ input-token speedup, or $1.4\times$ on average.
These matching points are measured along the full 100B-token learning-rate schedule.

\Needspace{5\baselineskip}
The adaptations add only 3.15M parameters, increasing the language backbone from 752.39M to 755.54M, or 0.42\%.
With this small parameter overhead, loop reuse consistently improves the pretrained model across all 11 domains.

\FloatBarrier

\ %
\section{Discussion}
\label{sec:discussion}

\para{Choosing among MoLT family architectures}
MoLT-R, MoLT-KV, and U-MoLT target distinct regimes across expressivity, efficiency, and backbone structure.
\Cref{tab:molt_variants} summarizes their mechanisms, inference profiles, and intended use cases.

MoLT-R prioritizes expressivity over single-traversal inference.
It keeps the residual stream across loops instead of passing earlier loops only through the normalized loop output.
Without adding parameters, MoLA directly reads current-loop and preceding-loop local KV at the same token position.
In \cref{tab:last10_accuracy,tab:last10_ppl}, MoLT-R reaches the highest 700M aggregate scores, with 59.2\% mean 10-task accuracy and 14.84 mean 10-domain PPL.
It also achieves the lowest validation CE among plain 12-layer stacks in \cref{tab:components}.
Our preliminary observations further indicate that its gain over plain residual feedback grows with effective computational depth, where information dilution in a single residual stream becomes more severe.
MoLT-R therefore fits expressivity-critical settings that can afford fixed $N$-loop decoding and $N$ sequence KV caches.

MoLT-KV prioritizes decoding efficiency with minimal architectural inductive bias.
Embedding restart and strict-prefix memory KV remove same-position cross-loop dependence.
Each decoding token can then skip preceding-loop traversals and run one traversal of the shared stack under MoL decoding.
Relative to fixed $N$-loop decoding, this cuts per-step decoding traversals and persistent KV storage by a factor of $N$.
At $N=2$, MoL-Hybrid decoding reduces total S-size MoLT-KV latency by nearly half in \cref{tab:decoding_quality_latency,tab:decoding_phase_latency}.
Meanwhile, reading cached prefix memory KV across the shared $L$-layer stack preserves multi-loop computation over the prefix, as analyzed in \cref{sec:prefix_stabilization}.
Because MoL-Hybrid-2 still runs $2L$ full-prompt prefill blocks in MoLT-KV, this variant fits generation-heavy workloads where decoding dominates latency, matching U-MoLT's decoding speed.
It also fits standard or hybrid backbones where additional architectural inductive bias is undesired.

U-MoLT builds on loop KV by adding an architectural inductive bias tailored to autoregressive language modeling.
It separates memory construction in a looped MoLT Encoder from prediction readout in a causal, cross-only MoLT Decoder.
By \cref{prop:decoder_omission}, U-MoLT omits unused decoder traversals during prefill and intermediate loops.
With an equal split $L_E=L_D=L/2$, MoL-Hybrid-2 removes the $2\times$ prefill block depth of MoLT-KV.
As shown in \cref{sec:cost}, it matches a non-looped $L$-layer backbone in full-prompt prefill block depth, quadratic prefill attention, decoding traversals, and persistent KV storage.
With independent decoder KV projections, U-MoLT also lowers validation CE relative to MoLT-KV across all matched settings in \cref{tab:components}.
U-MoLT is therefore our default choice for general-purpose pure-attention language models.

\begin{table}[htbp]
\centering
\caption{\textbf{Quick-reference comparison of the three MoLT variants.}
Costs compare an $L$-layer stack with $N$ loops, with $L_E+L_D=L$ in U-MoLT, against a non-looped $L$-layer backbone.
Inference profile lists the schedule, full-prompt prefill blocks, per-step decode blocks, and persistent KV storage.
Under MoL-Hybrid decoding with $k\in\{1,2\}$ prefill loops, persistent KV counts cached prefix KV across the shared stack and, for U-MoLT, the decoder.
U-MoLT prefill additionally runs $L_D$ single-token decoder blocks at the final prompt position.}
\label{tab:molt_variants}
\footnotesize
\setlength{\tabcolsep}{3.5pt}
\renewcommand{\arraystretch}{1.2}
\begin{tabularx}{\linewidth}{@{}
  >{\raggedright\arraybackslash}p{0.098\linewidth}
  >{\raggedright\arraybackslash}p{0.148\linewidth}
  >{\raggedright\arraybackslash}p{0.232\linewidth}
  >{\raggedright\arraybackslash}p{0.200\linewidth}
  >{\raggedright\arraybackslash}X
@{}}
\toprule
\textbf{Variant} & \textbf{Primary focus} & \textbf{Mechanism \& layout} & \textbf{Inference profile} & \textbf{Best suited for} \\
\midrule
\textbf{MoLT-R}
  & Maximum expressivity
  & Plain $L$-layer stack, loop residual, current- \& preceding-loop local KV
  & Fixed $N$-loop\newline $NL$ prefill, $NL$ decode\newline $N{\times}$ persistent KV
  & Expressivity-critical or deeper $N$-loop regimes with zero added parameters \\
\addlinespace[3pt]
\textbf{MoLT-KV}
  & Plain-stack efficiency
  & Plain $L$-layer stack, embedding restart, loop KV, current-loop local KV
  & MoL-Hybrid, $k{\in}\{1,2\}$\newline $kL$ prefill, $L$ decode\newline $1{\times}$ persistent KV
  & Generation-heavy workloads or plain backbones with less inductive bias \\
\addlinespace[3pt]
\textbf{U-MoLT}
  & Balanced efficiency \& quality
  & $L_E{+}L_D$ split, loop KV encoder, cross-only decoder with per-layer KV
  & MoL-Hybrid, $k{\in}\{1,2\}$\newline $kL_E$ prefill, $L$ decode\newline $1{\times}$ persistent KV
  & General-purpose pure-attention LLMs needing matched prefill and decode cost \\
\bottomrule
\end{tabularx}
\end{table}

\para{Representational decoupling in U-MoLT}
A natural question is why U-MoLT outperforms MoLT-KV in validation CE even though it loops only half the layers.
We attribute this gain to decoupling two conflicting roles in deep autoregressive layers.
In a standard stack, deep layers near the prediction head face two competing pressures.
First, for \emph{prediction}, hidden states near the output must progressively collapse toward the next-token distribution $p(x_{t+1}\mid x_{\leq t})$.
This process discards or rewrites fine-grained semantic details of the current token.
Second, for \emph{memory}, self-attention KV are projected directly from each layer's hidden states.
Those same deep layers must therefore preserve clean historical context for future tokens at $t'>t$.
MoLT-KV sharpens this conflict because its normalized loop output $z^r$ feeds the vocabulary projection while also supplying memory KV to the next loop.

U-MoLT physically isolates these two responsibilities.
The normalized encoder output $z^r$ sits midway through the network rather than directly beneath the prediction head.
It can therefore act as a high-fidelity shared memory without collapsing toward next-token logits.
In contrast, hidden states in the cross-only MoLT Decoder never supply KV for future tokens and never feed back into the encoder.
The decoder residual stream can therefore specialize toward predicting $x_{t+1}$ without contaminating the memory read by later tokens.
Even at the deepest decoder layers, independent decoder KV projections can still read clean context from $z^r$ right before the final prediction.
Our ablations in \cref{tab:components} are consistent with this view.
Across all 6 matched U-MoLT triplets, decoder self-attention under the \emph{self} setting raises CE by 0.014--0.021 relative to independent per-layer projections under \emph{re-proj.}
Independent projections also beat \emph{shared} projections by 0.009--0.018 CE, showing that decoder layers benefit from separate views of the same normalized encoder output.
In \cref{tab:qwen_ct}, however, this cross-only decoder design also depends on the backbone's sequence layers.
On Qwen3.5, forcing decoder GDN layers to read only encoder-projected memory in U-MoLT-CT2 lags U-MoLT-CT1 by 0.003 mean CE across the 11 domains, while still not achieving exact single-position prefill omission (\cref{sec:qwen_ct}).
On hybrid backbones with non-attention sequence mixers, retaining decoder-state recurrence in hybrid U-MoLT adaptations or using plain MoLT stacks therefore remains an important alternative.

\para{Beyond two stages: Encoder--Neck--Decoder architectures}
U-MoLT splits a looped Transformer into two stages, using a looped encoder for memory construction and a cross-only decoder for prediction readout.
A natural generalization is a three-stage Encoder--Neck--Decoder architecture.
In this layout, a non-looped encoder extracts foundational context, a looped middle \emph{neck} performs iterative cross-loop refinement, and a cross-only decoder handles prediction readout.
Informally, this forms a ``giraffe'' network whose recurrent depth is elongated in the middle neck while keeping the encoder and decoder non-looped.
In our preliminary experiments extending U-MoLT to this three-stage layout, looping the middle neck achieved the strongest validation results.
This points toward an even cleaner separation among input encoding, iterative refinement, and prediction readout.
We leave a systematic study of this ``giraffe'' architecture, including stage partitioning and inference schedules, to future work.

\para{Scaling to larger models}
Computational constraints have limited our scaling experiments to 700M and 1.5B dense models and 3B MoE models.
How the gains from loop reuse evolve with larger model capacity and training budgets remains an open question.
A broader scaling study could compare allocating additional computation to larger models or to additional loops, measuring prediction quality alongside training and inference cost.
Such experiments would help determine how to balance parameter capacity and shared-weight loop computation as MoLT scales up.

\Needspace{9\baselineskip}
\section{Conclusion}
In this paper, we present Mixture-of-Loops Transformer (\name{}), which uses keys and values (KV) as memory to selectively reuse computation across depths and loops.
Through Mixture-of-Loops Attention (\attn{}), \name{} spans MoLT-R under loop residual for expressivity and MoLT-KV under loop KV for plain-stack efficiency.
Under loop KV, embedding restart and strict-prefix memory KV remove same-position cross-loop dependence.
Trained via loop forcing, this enables single-traversal MoL decoding beyond the training loop horizon, while MoL-Hybrid decoding adds parallel prefill.
To balance efficiency and quality, U-\name{} further separates memory construction from prediction readout and omits unused decoder traversals at inference.
Experiments on 700M and 1.5B dense models and 3B MoE models trained with the OLMo3 recipe show consistent gains in perplexity and downstream accuracy.
Continued pretraining on Qwen3.5-0.8B-Base confirms effective transfer to pretrained backbones, and MoL-Hybrid decoding achieves substantial latency reductions over fixed-loop decoding.
These results suggest that explicit KV reuse across depths and loops is a practical and effective primitive for scaling computational depth in looped Transformers.
We will release the full implementation of \name{}, and we hope it will serve as a foundation for building stronger looped large language models in the open-source community.
\name{} integrates naturally with dense, MoE, and hybrid backbones.
Beyond language modeling, it can be readily extended to multimodal intelligence, visual understanding, and world models, where iterative Transformer computation is increasingly adopted.
We believe that principled information reuse across depths and loops will bring broad and lasting benefits across these diverse domains.
\setlength{\bibsep}{3pt}
\begingroup
\interlinepenalty=10000
\bibliographystyle{molt_plainnat}
\bibliography{molt}
\endgroup
\clearpage
\appendix

\begin{figure}[!t]
\centering
\includegraphics[width=\textwidth]{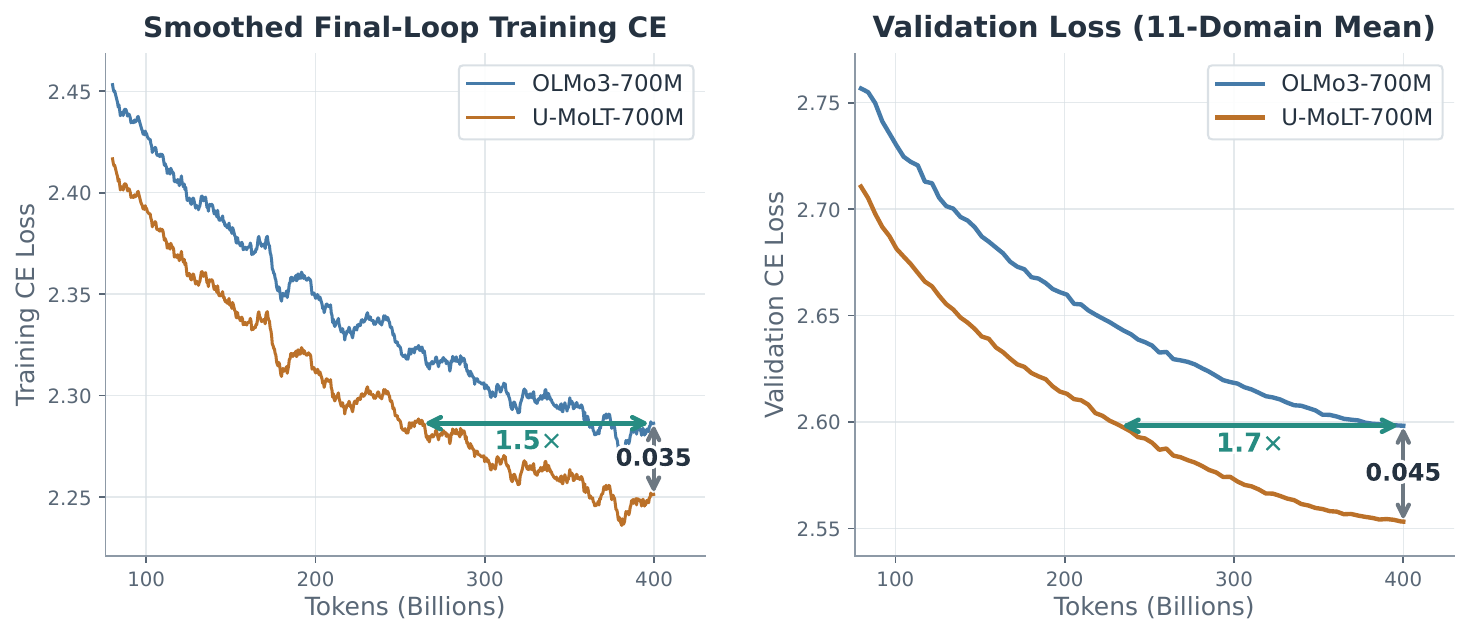}
\caption{\textbf{Training and validation token efficiency at 700M scale from 80B to 400B tokens.}
The left panel plots smoothed final-loop training CE, sampled every 100 steps through step 95,300 with a 20-point moving average.
The right panel plots the separate 11-domain mean validation CE curve.
Teal arrows trace horizontally from OLMo3's final plotted CE to the matching U-MoLT curve, and gray arrows show the final CE gap.
U-MoLT achieves a $1.5\times$ input-token speedup on smoothed final-loop training CE with a final gap of 0.035.
On 11-domain mean validation CE, U-MoLT reaches OLMo3's 400B loss at 232.8B tokens, yielding a $1.7\times$ speedup and a final gap of 0.045.}
\label{fig:large_loss_speedup}
\end{figure}

\subsection{Per-Domain Validation Trajectories}
\label{sec:large_val_domains}
\begin{figure}[!t]
\centering
\includegraphics[width=\textwidth]{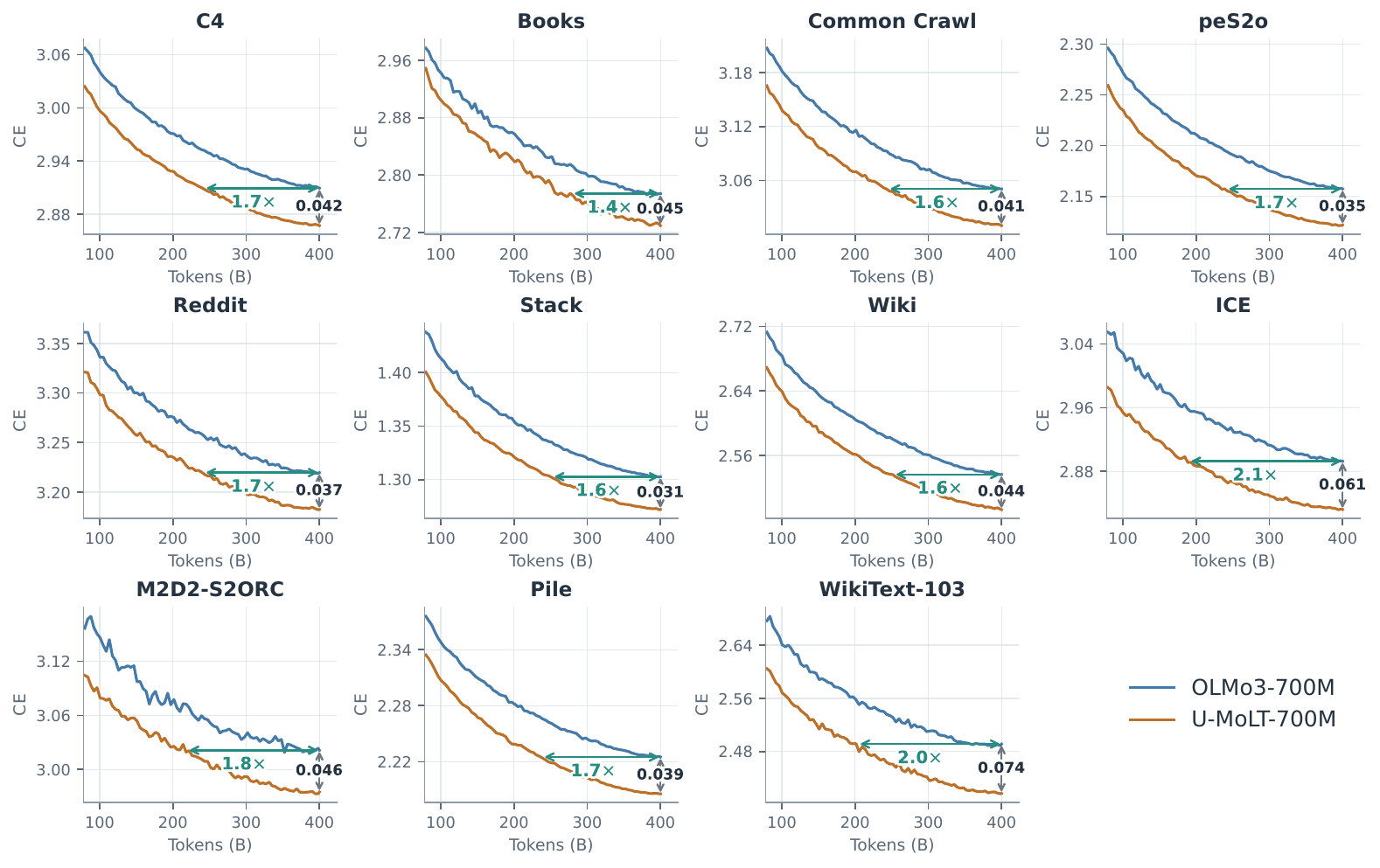}
\caption{\textbf{Per-domain validation CE trajectories at 700M scale from 80B to 400B tokens.}
Each panel shows validation CE for OLMo3-700M and U-MoLT-700M, where lower is better.
Teal arrows mark where U-MoLT reaches OLMo3's final 400B CE via linear interpolation, and gray arrows show the final CE gap.
Annotated factors report the resulting input-token speedup, ranging from $1.4\times$ to $2.1\times$ across domains.}
\label{fig:large_val_ce}
\end{figure}

\section{700M/400B Convergence and Scaling Study}
\label{sec:large_details}
This appendix details the paired 700M/400B convergence and compute study in \cref{sec:large} across 11 periodic-evaluation domains (including Dolma Wiki).
Because it uses a separate 400B data split from \cref{tab:last10_accuracy,tab:last10_ppl}, absolute CE values differ slightly from $\ln(\text{PPL})$ in \cref{tab:last10_ppl}, with consistent relative gains across shared domains.

\begin{table}[!t]
\centering
\caption{\textbf{Per-domain validation CE and token speedup at 700M scale and 400B tokens.}
Paired 11-domain run on a separate 400B data split from \cref{tab:last10_ppl}.
Lower CE is better, and $\Delta\text{CE}$ and $\Delta\text{PPL}=100[\exp(\Delta\text{CE})-1]$ use unrounded values, giving the relative change in geometric mean perplexity on the equal-domain mean row.
Match tokens is where U-MoLT reaches OLMo3's 400B CE.
Speedup is $400.0\text{B}/\text{Match tokens}$.}
\label{tab:large_domains}
\small
\begin{tabular*}{\textwidth}{@{\extracolsep{\fill}}lrrrrrr@{}}
\toprule
Domain & OLMo3 CE & U-MoLT CE & $\Delta$ CE & $\Delta$ PPL (\%) & Match tokens (B) & Speedup \\
\midrule
C4 & 2.909 & 2.867 & $-0.042$ & $-4.13\%$ & 240.4 & $1.7\times$ \\
Books & 2.774 & 2.729 & $-0.045$ & $-4.38\%$ & 277.1 & $1.4\times$ \\
Common Crawl & 3.051 & 3.010 & $-0.041$ & $-4.00\%$ & 242.6 & $1.6\times$ \\
peS2o & 2.157 & 2.122 & $-0.035$ & $-3.49\%$ & 239.2 & $1.7\times$ \\
Reddit & 3.220 & 3.182 & $-0.037$ & $-3.66\%$ & 240.0 & $1.7\times$ \\
Stack & 1.303 & 1.272 & $-0.031$ & $-3.01\%$ & 249.9 & $1.6\times$ \\
Wiki & 2.536 & 2.493 & $-0.044$ & $-4.27\%$ & 250.5 & $1.6\times$ \\
ICE & 2.893 & 2.832 & $-0.061$ & $-5.87\%$ & 187.9 & $2.1\times$ \\
M2D2-S2ORC & 3.021 & 2.975 & $-0.046$ & $-4.52\%$ & 217.1 & $1.8\times$ \\
Pile & 2.225 & 2.186 & $-0.039$ & $-3.87\%$ & 236.5 & $1.7\times$ \\
WikiText-103 & 2.491 & 2.417 & $-0.074$ & $-7.15\%$ & 201.6 & $2.0\times$ \\
\midrule
Equal-domain mean & 2.598 & \textbf{2.553} & $-0.045$ & $-4.40\%$ & 232.8 & $\bm{1.7\times}$ \\
\bottomrule
\end{tabular*}
\end{table}

\begin{table}[!t]
\centering
\caption{\textbf{Configuration, training cost, and token efficiency for the 700M models trained on 400B tokens.}
Wall-clock runtime and GPU-hours cover the complete 400B-token training run.}
\label{tab:large_cost_summary}
\small
\setlength{\tabcolsep}{4pt}
\begin{tabular*}{\textwidth}{@{\extracolsep{\fill}}lrrr@{}}
\toprule
Quantity & OLMo3 baseline & U-MoLT & Ratio / $\Delta$ \\
\midrule
Unique trainable parameters & 700.53M & 711.02M & $+1.50\%$ \\
Backbone organization & 38L (1 loop) & 19L + 19L (2 loops) & shared weights \\
Block execution (train / eval) & 38 / 38 & 76 / 57 & $2.00\times$ / $1.50\times$ \\
Theoretical training FLOPs (PFLOPs) & $2.446\times 10^{6}$ & $4.918\times 10^{6}$ & $2.01\times$ \\
Total wall-clock runtime (hours) & 110.1 & 219.2 & $1.99\times$ \\
Total compute (H20 GPU-hours, 64 GPUs) & 7,049 & 14,026 & $1.99\times$ \\
Peak active GPU memory (GiB) & 43.6 & 58.4 & $1.34\times$ \\
Tokens to match 400B baseline CE & 400.0B & \textbf{232.8B} ($1.7\times$) & $-41.8\%$ \\
\bottomrule
\end{tabular*}
\end{table}

\subsection{Training Convergence and Compute Cost}
\label{sec:large_cost_details}
\suppressfloats[t]
Both models are trained from scratch on a deterministic 400B-token split of \code{OLMo-mix-0625-official} with 4,096-token contexts and a 4.19M-token global batch over 95,368 steps.
They use the Dolma2 tokenizer with a padded vocabulary size of 100,352 and tied embeddings.
Optimization uses AdamW with $\beta_1=0.9$, $\beta_2=0.95$, weight decay 0.1 excluding embeddings, and gradient clipping 1.0.
The learning rate warms up over 2,000 steps to $3\times 10^{-4}$ and follows cosine decay to $3\times 10^{-5}$.
The OLMo3 baseline is a 38-layer dense Transformer with width 1,024, 16 query heads, 8 KV heads~\citep{ainslie2023gqa}, QK normalization, and 700.53M unique parameters.
U-MoLT partitions the 38 blocks into a 19-layer encoder and a 19-layer decoder, totaling 711.02M parameters, or $+1.50\%$.
It runs $N=2$ shared-weight loops with sparse MoLA placement at even encoder layers and full-history current-loop depth KV.
Because validation only uses the final-loop predictions, U-MoLT omits the first-loop decoder during evaluation and executes 57 blocks per token versus 38 in the baseline.

\begin{figure}[!t]
\centering
\includegraphics[width=\textwidth]{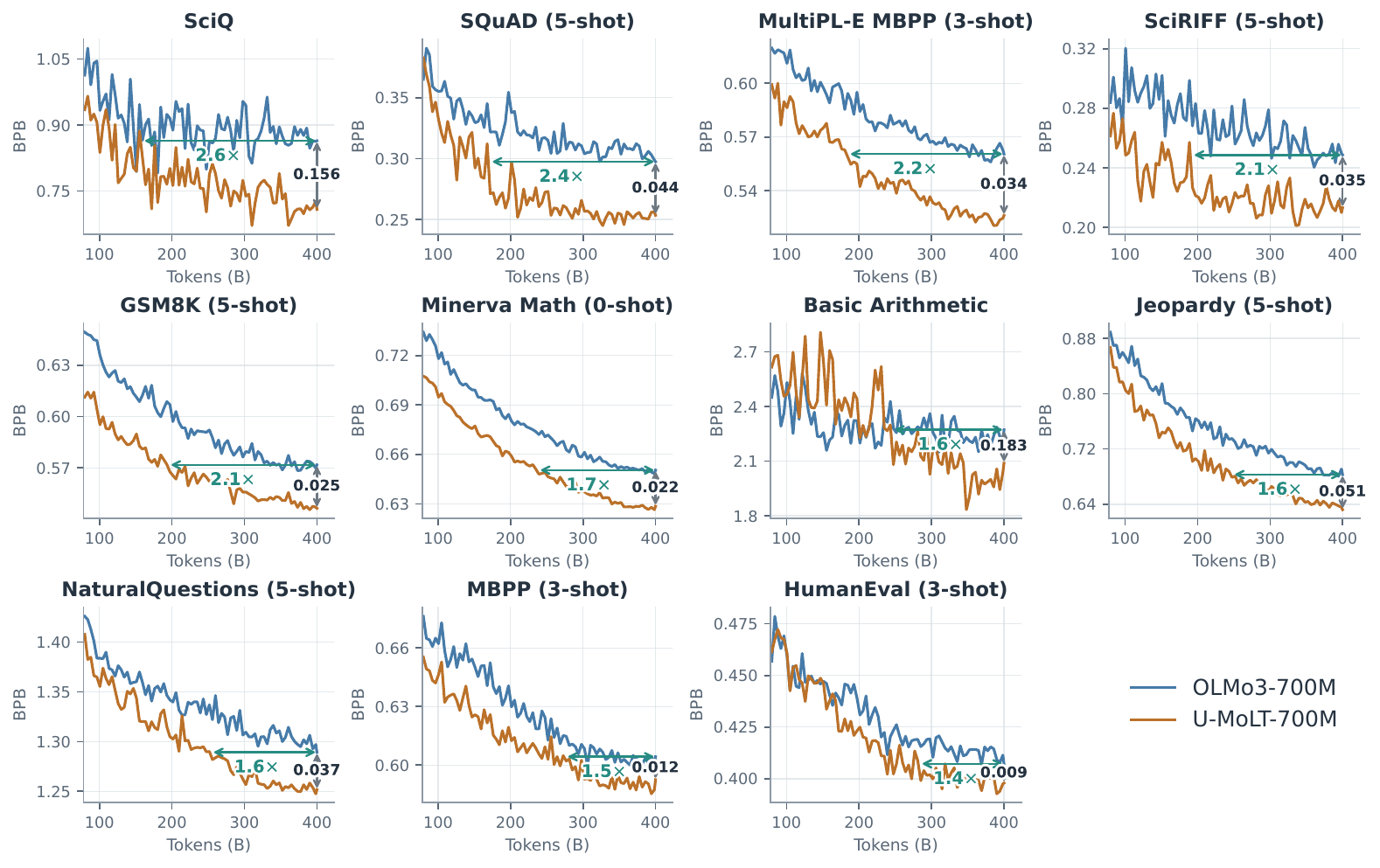}
\caption{\textbf{Downstream answer BPB trajectories at 700M scale from 80B to 400B tokens.}
Each panel shows gold-answer BPB for OLMo3-700M and U-MoLT-700M across 11 representative benchmarks, where lower is better.
Teal arrows mark where U-MoLT reaches OLMo3's final 400B BPB via linear interpolation, with input-token speedups from $1.4\times$ to $2.6\times$, and gray arrows show the final BPB gap.
Across all 55 answer-BPB evaluations in the suite, U-MoLT improves 44 evaluations, reducing mean BPB from 1.094 to 1.075 with a $1.6\times$ token speedup.}
\label{fig:large_downstream_bpb}
\end{figure}

\Cref{tab:large_cost_summary,fig:large_loss_speedup} summarize the compute cost and token efficiency of the 700M/400B comparison.
With only $+1.50\%$ additional parameters, U-MoLT achieves a $1.5\times$ token speedup on smoothed final-loop training CE with a final gap of 0.035.
On 11-domain mean validation CE, it matches OLMo3's 400B endpoint at 232.8B tokens, yielding a $1.7\times$ token speedup, or 41.8\% fewer tokens.
In total compute, 2-loop training uses $2.01\times$ theoretical FLOPs, $1.99\times$ wall-clock runtime, and $1.34\times$ peak active GPU memory.
Because U-MoLT reaches the baseline's final FLOP budget at 198.9B tokens, midway through the 400B cosine schedule, an exact equal-compute comparison requires a dedicated run decayed to completion at that budget.

\Cref{fig:large_val_ce,tab:large_domains} report validation CE across all 11 domains.
U-MoLT leads across all 11 domains from 12.6B tokens through 400B tokens.
At 400B tokens, equal-domain mean CE drops by 0.045, from 2.598 to 2.553, corresponding to a $4.40\%$ reduction in geometric mean perplexity across the 11 domains.
Per-domain perplexity reductions range from $3.01\%$ on Stack to $7.15\%$ on WikiText-103.
Across domains, matching OLMo3's final CE yields a $1.4$--$2.1\times$ input-token speedup, or $1.7\times$ overall, reaching $2.0\times$ on WikiText-103 and $2.1\times$ on ICE.

\FloatBarrier
\subsection{Downstream Answer-Likelihood Trajectories}
\label{sec:large_downstream_details}
On these paired 700M/400B runs, \cref{fig:large_downstream_bpb} tracks gold-answer bits per byte (BPB), where lower is better, across 11 representative science QA, reading-comprehension, math, and code benchmarks, complementing \cref{tab:last10_accuracy}.
Across all 55 answer-BPB evaluations in the OLMo3 suite, U-MoLT improves 44 evaluations and reduces the overall mean from 1.094 to 1.075, matching OLMo3's 400B endpoint at 243.3B tokens for a $1.6\times$ token speedup.
Notable token speedups reach $2.6\times$ on SciQ, $2.4\times$ on SQuAD, $2.2\times$ on MultiPL-E MBPP, $2.1\times$ on SciRIFF and GSM8K, and $1.7\times$ on Minerva Math.
Because BPB evaluates teacher-forced log-likelihood on reference solutions, it measures answer probability rather than free-form generation pass rates.

\FloatBarrier
\section{Analyzing Loop KV: Recurrence, Loop Forcing, and Efficiency}
\label{sec:analysis}

Why can loop KV break the sequential-loop decoding bottleneck while preserving multi-loop computation over the prefix, whereas loop residual cannot?
We answer this question for the all-layer MoLA stacks of MoLT-KV and U-MoLT in three steps.
First, we derive the strictly triangular recurrence of loop KV and verify autoregressive causality (\cref{sec:prefix_stabilization}).
This establishes single-traversal equivalence for MoL and MoL-Hybrid decoding and quantifies the resulting multi-loop path depth over the prefix.
Second, we explain how loop forcing trains this token-ordered recurrence in parallel and how 2-loop training prepares the model for recursively generated prefix memory KV (\cref{sec:loop_forcing}).
Third, we prove decoder omission for U-MoLT and quantify traversal, KV-storage, and per-traversal computational costs (\cref{sec:cost}).

\subsection{Strictly Triangular Recurrence and Single-Traversal Decoding}
\label{sec:prefix_stabilization}
\para{Strictly triangular recurrence}
Under loop residual, token $t$ in feedback loop $r>1$ depends on its own preceding-loop output $z^{r-1}_t$.
In MoLT-R, it also reads preceding-loop local KV $\mathcal D_{\mathrm{prev}}$ at position $t$.
This same-position cross-loop dependence forces inference to run all $N$ loops sequentially before producing a token's final output.

Under loop KV, embedding restart fixes the layer-0 input at $X^{r,0}_t=e(x_t)$, and strict-prefix memory reads restrict $\mathcal S_{\mathrm{mem}}$ to $z^{r-1}_{<t}$.
Current-loop depth KV $\mathcal D_{\mathrm{cur}}$ come only from layers $k\leq\ell$ at position $t$, so they introduce no cross-loop dependence.
Layer projections, RoPE, normalization, and feed-forward updates are also position-wise.
For fixed inputs and deterministic evaluation, every feedback loop therefore reduces to the strictly triangular map
\begin{equation}
 z^r_t=F_t(x_t,z^{r-1}_{<t}),\qquad r\geq2.
 \label{eq:triangular}
\end{equation}
The index $t$ records the token's RoPE position and prefix length.
Parameters and depth-source selection are shared across loops without loop-index dependence.
Together with embedding restart from $e(x_t)$, this makes $F_t$ identical for all $r\geq2$.
Each evaluation of $F_t$ runs one traversal of the shared stack with MoLA from $e(x_t)$, followed by output normalization.
At $t=0$, strict-prefix sequence KV are empty, so including own KV keeps every layer's KV set nonempty.

Loop~1 reads current-loop sequence KV at $j<t$ and current-loop depth KV at $k\leq\ell$ of position $t$.
Its raw and normalized outputs therefore depend only on $x_{\leq t}$.
Inside each evaluation of $F_t$ in \cref{eq:triangular}, every layer reads only strictly earlier positions $j<t$ and current-position states at $t$.
Induction over loops keeps every raw and normalized loop output dependent only on $x_{\leq t}$ under fixed-loop decoding.
Induction over token positions gives the same property for MoL decoding, and for MoL-Hybrid decoding after its fixed-loop prefill.
With MoLT-KV's position-wise prediction head and U-MoLT's causal cross-only decoder over $j\leq t$, all three schedules preserve autoregressive causality.

\para{Finite-prefix stabilization and empty-cache MoL decoding}
Because \cref{eq:triangular} has no diagonal dependence across loops, a sufficiently long unroll stabilizes from earlier positions to later ones.
LCKV proves finite-iteration equivalence for top-layer feedback under a similar triangular dependency~\citep{wu2024lckv}.
Here the same induction extends to per-layer memory KV projections from the loop output, current-loop depth KV, and a distinct first-loop initialization.
It also carries over to U-MoLT's decoder readout and MoL-Hybrid's parallel prefill below.
\begin{proposition}[Finite-prefix stabilization]
\label{prop:stability}
For fixed inputs and shared deterministic maps in \cref{eq:triangular} with own KV included in $\mathcal D_{\mathrm{cur}}$, there is a unique stabilized solution
$z^\star_t=F_t(x_t,z^\star_{<t})$.
Starting from any first-loop states $z^1$, position $t$ equals $z^\star_t$ for every $r\geq t+2$.
\end{proposition}
\begin{proof}
The solution is constructed uniquely in token order.
At $t=0$, $F_0$ has no memory argument, so $z^r_0=F_0(x_0)=z^\star_0$ for all $r\geq2$, regardless of $z^1_0$.
Suppose the claim holds for positions $j<t$.
For $r\geq t+2$, the preceding loop satisfies $r-1\geq j+2$ for every $j<t$.
Hence $z^{r-1}_{<t}=z^\star_{<t}$, and substitution into \cref{eq:triangular} gives $z^r_t=z^\star_t$.
No contraction assumption is needed because the dependency graph is strictly triangular.
\end{proof}

MoL decoding implements this token-ordered recurrence directly from empty caches.
\begin{corollary}[Token-ordered recurrence]
\label{cor:token_ordered}
Assume the conditions of \cref{prop:stability}, together with the deterministic shared stack and readouts in \cref{eq:boundary,eq:decoder,eq:forms}.
Processing a fixed input sequence from empty caches by evaluating $F_0,F_1,\ldots,F_{T-1}$ once in token order produces the stabilized solution $z^\star$.
At every position $t$, the raw outputs and next-token distributions match every unroll of depth $r\geq t+2$.
Across the length-$T$ sequence, they match every unroll of depth $r\geq T+1$.
\end{corollary}
\begin{proof}
By induction on $t$, evaluating $F_0,F_1,\ldots,F_{T-1}$ from empty caches yields $z^\star_t$ at every position.
For any $r\geq t+2$, \cref{prop:stability} gives $z^{r-1}_{<t}=z^\star_{<t}$.
Both executions therefore enter the shared stack at position $t$ with the same embedding $e(x_t)$ and normalized prefix outputs $z^\star_{<t}$.
Induction over layers preserves all current-loop depth KV and the raw output $X^{r,L}_t$ or $X^{r,L_E}_t$.
In MoLT-KV, the prediction head receives the same raw output, so the next-token distributions agree.
In U-MoLT, every position $j\leq t$ satisfies $r\geq j+2$, so $z^r_{\leq t}=z^\star_{\leq t}$.
The decoder starts from the same raw encoder output and projects the same decoder KV from $z^\star_{\leq t}$, so the next-token distributions also agree.
\end{proof}
Because every position yields the same next-token distribution, token-by-token induction extends this equivalence to autoregressive generation under greedy decoding or matched sampling draws.
At each step $t$, empty-cache MoL decoding matches any unroll of depth $r\geq t+2$ evaluated on the generated prefix $x_{\leq t}$.

\para{Parallel prefill and conditional stabilization under MoL-Hybrid decoding}
Empty-cache MoL decoding recovers the full-sequence stabilized solution $z^\star$ (\cref{cor:token_ordered}), but it processes prompt tokens sequentially.
MoL-Hybrid decoding keeps parallel prefill over a length-$P$ prompt by initializing prompt memory KV from parallel $k$-loop outputs $z^k_{<P}$, with $k\in\{1,2\}$.
In U-MoLT, prompt decoder KV are also projected from $z^k_{<P}$.
After predicting the first continuation token $x_P$ at position $P-1$, MoL-Hybrid decoding applies $F_t$ once per continuation token $t=P+m$ for $m\geq 0$:
\begin{equation}
 z^{\mathrm{hyb}}_t=F_t\!\left(x_t,\left[z^k_{<P},z^{\mathrm{hyb}}_{P:t-1}\right]\right),\qquad t\geq P.
 \label{eq:hybrid_triangular}
\end{equation}
Because prompt states stay fixed at $z^k_{<P}$ instead of stabilizing to $z^\star_{<P}$, the full-sequence guarantee of \cref{cor:token_ordered} applies only to empty-cache MoL decoding.
However, \cref{eq:hybrid_triangular} preserves the same strictly triangular structure over continuation indices $m=t-P\geq 0$.
Parallel prefill therefore does not cap continuation depth at $k$ loops.
To formalize this, we compare \cref{eq:hybrid_triangular} with a multi-loop unroll that holds prompt outputs fixed at $z^k_{<P}$ and repeatedly updates only the continuation.
\begin{corollary}[Prompt-conditioned stabilization]
\label{cor:hybrid_stabilization}
Fix a length-$P$ prompt $x_{<P}$ with frozen prompt outputs $z^k_{<P}$ from parallel $k$-loop prefill, and consider fixed continuation inputs $x_P,\ldots,x_{T-1}$ indexed by $m=t-P\geq 0$.
Under the assumptions of \cref{cor:token_ordered}, applying \cref{eq:hybrid_triangular} once per continuation token yields a unique stabilized continuation $z^{\mathrm{hyb}}_{P:T-1}$.
Consider any fixed-prompt continuation unroll defined for $s\geq 2$ by $\tilde z^s_t=F_t(x_t,[z^k_{<P},\tilde z^{s-1}_{P:t-1}])$.
It starts from arbitrary initial continuation states $\tilde z^1_{P:T-1}$ while keeping $z^k_{<P}$ frozen.
In U-MoLT, both schedules use the same frozen prompt decoder KV projected from $z^k_{<P}$.
At every continuation position $t=P+m$, the raw outputs and next-token distributions of \cref{eq:hybrid_triangular} match every such continuation unroll of depth $s\geq m+2$.
\end{corollary}
\begin{proof}
Define the shifted maps $\widetilde F_m(x_{P+m},u_{<m})=F_{P+m}(x_{P+m},[z^k_{<P},u_{<m}])$ for $m\geq 0$.
Because $z^k_{<P}$ is fixed across loops, $\{\widetilde F_m\}_{m\geq 0}$ has the same strictly triangular form as \cref{eq:triangular} over $m\geq 0$.
Applying \cref{prop:stability} to $\{\widetilde F_m\}_{m\geq 0}$ yields the unique stabilized continuation $z^{\mathrm{hyb}}_{P:T-1}$, with $\tilde z^s_t=z^{\mathrm{hyb}}_t$ for all $s\geq m+2$.
For any $s\geq m+2$, every earlier continuation position $j$ with $P\leq j<t$ satisfies $s-1\geq (j-P)+2$.
Both schedules therefore enter the shared stack at $t=P+m$ with the same $e(x_t)$, frozen $z^k_{<P}$, and continuation prefix $z^{\mathrm{hyb}}_{P:t-1}$.
Induction over layers therefore preserves all current-loop depth KV and the raw loop output, so MoLT-KV next-token distributions agree.
In U-MoLT, prompt decoder KV at $j<P$ come from the same frozen $z^k_{<P}$, and every continuation position $P\leq j\leq t$ satisfies $s\geq (j-P)+2$.
Hence $\tilde z^s_j=z^{\mathrm{hyb}}_j$ for $P\leq j\leq t$.
The decoder therefore starts from the same raw encoder output and reads the same decoder KV, giving identical next-token distributions.
\end{proof}
The same token-by-token induction extends this equivalence to autoregressive generation under greedy decoding or matched sampling draws.
With continuation recursion preserved, the two prefill depths trade prompt representation quality against prefill cost.
MoL-Hybrid-1 minimizes prefill latency and projects prompt memory KV from $z^1_{<P}$, matching the memory source seen during 2-loop training.
However, its prompt states stay at single-loop depth, and its first continuation prediction at $P-1$ remains at loop-1 quality.
In U-MoLT, its prompt decoder KV are also projected from $z^1_{<P}$ throughout decoding.
MoL-Hybrid-2 matches fixed 2-loop prediction at $P-1$ and retains 2-loop prompt depth through $z^2_{<P}$.
It projects prompt memory KV, and prompt decoder KV in U-MoLT, from $z^2_{<P}$ for all continuation tokens.

\para{Extension to sparse feedback placement}
The stabilization results in \cref{prop:stability,cor:token_ordered,cor:hybrid_stabilization} also extend to sparse feedback placement, such as the alternating-layer MoLT-KV checkpoint evaluated in \cref{sec:execution_results}.
Each non-feedback layer reads causal current-loop sequence KV at $j\leq t$ and current-loop depth KV at $t$ when enabled, without directly reading preceding-loop KV.
The position induction in \cref{prop:stability,cor:token_ordered} therefore tracks every non-feedback layer's current-loop sequence KV cache together with the normalized loop outputs $z^r_t$.
At $r\geq t+2$, both $z^{r-1}_{<t}$ and each non-feedback layer's current-loop sequence KV at $j<t$ have stabilized.
Layer induction at position $t$ then determines each non-feedback layer's current-loop sequence KV at $t$, the raw output, and $z^\star_t$.
For \cref{cor:hybrid_stabilization}, both schedules also freeze and reuse the same loop-$k$ current-loop sequence KV at every non-feedback layer over $j<P$.
MoL-Hybrid decoding then matches every such unroll of depth $s\geq m+2$.

\para{Multi-loop computation over the prefix}
The stabilization bounds $r\geq t+2$ and $s\geq m+2$ reflect recursive memory propagation across tokens rather than repeated traversals of the same token.
In a non-looped $L$-layer Transformer, layer $\ell$ reads sequence KV only from layer $\ell$ of earlier tokens.
Because information never feeds from deeper layers back to shallower ones, any causal path traverses at most $L$ blocks.
For the all-layer stacks analyzed above, every shared-stack layer starting from layer~0 reads per-layer memory KV projected from completed normalized loop outputs at $j<t$.
In fixed 2-loop execution, a causal path from an earlier token $x_j$ with $j<t$ traverses up to $2L$ blocks in MoLT-KV and $2L_E+L_D$ blocks in U-MoLT.
Even with an equal split $L_E=L_D=L/2$, U-MoLT reaches $3L/2$ blocks over prefix context, exceeding a non-looped $L$-layer Transformer.
Under MoL decoding, each token passes through the shared stack once from $e(x_t)$, yet each $z^\star_j$ builds on earlier stabilized outputs.
From every earlier token $j<t$, the longest causal path traverses $(t-j+1)L$ blocks in MoLT-KV or $(t-j+1)L_E+L_D$ blocks in U-MoLT.
Because $t-j+1\geq 2$, this maximum path depth is always at least the fixed 2-loop depth.
MoL-Hybrid continuation steps preserve the same cross-token recursion from frozen prompt outputs $z^k_{<P}$.

Because embedding restart removes same-token cross-loop dependence and strict-prefix memory excludes $j=t$, the current token traverses the shared stack only once along the path to its own output.
Loop KV therefore trades same-token looping for cross-token memory recursion.
Within each traversal, current-loop depth attention over $\mathcal D_{\mathrm{cur}}$ supplies same-position attention, and U-MoLT's cross-only decoder adds causal readout over normalized encoder outputs at $j\leq t$.
The empirical contribution of these components is measured in \cref{tab:components}.
A single traversal per token preserves multi-loop computation over the prefix through cross-token memory recursion.

\subsection{Parallel Training with Loop Forcing and Cross-Loop Alignment}
\label{sec:loop_forcing}
The preceding results identify the unroll computed by single-traversal decoding, leaving two practical questions.
First, how can we train $F_t$ without running a sequential recurrence across all $T$ token positions?
Second, how does training on 2 loops prepare the model to use recursively generated prefix memory KV, and how does this change in memory source affect continuation quality?

\para{Parallelizing feedback training with loop forcing}
Under loop residual, training and fixed-loop inference execute the same $N$-loop computation.
Under loop KV, single-traversal decoding instead evaluates the token-ordered recurrence $z^\star_t=F_t(x_t,z^\star_{<t})$.
Directly training this recurrence requires $T$ sequential steps across the sequence, which is prohibitive at long context lengths.
Loop forcing restores sequence parallelism in the same way teacher forcing parallelizes autoregressive training.
For each feedback loop $r\geq 2$, loop forcing conditions $F_t$ on the preceding loop's normalized outputs $z^{r-1}_{<t}$ rather than on the sequentially constructed prefix $z^\star_{<t}$.
Loop~1 computes $z^1$ in parallel using current-loop sequence KV.
Each subsequent loop evaluates $F_t$ in parallel across all positions.
The multi-loop loss in \cref{eq:loss} supervises every loop with positive weight.
Training therefore runs in $N$ parallel passes instead of $T$ sequential steps.
Loop forcing works with either open or detached cross-loop gradients.
Open gradients let later-loop losses also refine earlier loop outputs through memory KV.

\para{Cross-loop alignment and continuation quality}
The hardest extrapolation setting is $N=2$, where $F_t$ never sees its own outputs during training.
Analogous to exposure bias in autoregressive decoding, 2-loop training evaluates $F_t$ only on memory KV projected from loop-1 outputs $z^1_{<t}$.
Those outputs are produced with current-loop sequence KV rather than by $F_t$ itself.
During MoL decoding, $F_t$ instead reads prefix memory KV projected from its own earlier outputs.
Under MoL-Hybrid decoding with $k\in\{1,2\}$ prefill loops, $F_t$ reads frozen prompt memory KV projected from $z^k_{<P}$ together with continuation memory KV recursively appended from its own earlier outputs.
Without alignment across loops, recursive extrapolation over hundreds of continuation tokens could cause representation drift.
Three training and architectural choices help support this shift in memory source.

First, supervising next-token prediction at every loop through the shared prediction readout encourages $z^1$ and $z^2$ to align under 2-loop training.
In MoLT-KV, the shared vocabulary projection maps each supervised $z^r$ directly to next-token logits.
In U-MoLT, the shared cross-only decoder starts from $X^{r,L_E}$ and reads $z^r$ through independent decoder KV projections across all $L_D$ layers.
Both loops also run the same shared stack from $e(x_t)$.
Together, these shared components encourage $z^1$ and $z^2$ to remain compatible when memory KV projections trained on $z^1$ are applied to $z^2$ and later recursive outputs.
With detached gradients, loop-1 loss is the only loss that backpropagates into $z^1$, directly aligning it with the shared readout.
With open gradients, later-loop losses also backpropagate into $z^1$ through memory KV.
Positive loop-1 weight then prevents later-loop losses from reshaping $z^1$ solely as an internal feedback source.
In \cref{tab:loop_loss_weights}, raising the relative weight on loop~2 lowers loop-2 CE but raises loop-1 CE.
Loop-loss weights therefore trade loop-1 CE against loop-2 CE.

Second, embedding restart anchors each token's residual stream to $e(x_t)$, and single-traversal decoding runs the shared stack only once per token.
Even when memory propagates recursively across hundreds of earlier positions, earlier normalized loop outputs never overwrite the current token's residual stream.
Instead, they enter only through softmax attention over prefix memory KV, competing with current-loop depth KV along the current token's residual stream.
Unlike unrolling loop residual to depth $r\geq t+2$ at the same token, long continuations under MoL decoding therefore avoid compounding same-token residual drift away from $e(x_t)$.

Third, even without a contraction assumption, normalization keeps activations and attention logits uniformly bounded across arbitrarily long recursive histories.
For fixed parameters, output normalization bounds every normalized loop output and its projected memory values independently of position and unroll depth.
Starting from a bounded token embedding $e(x_t)$, each layer forms a softmax convex combination of sequence and depth values before its residual and feed-forward updates.
Across a single finite traversal, layer induction therefore keeps every layer state and its current-loop local values bounded regardless of sequence length.
In U-MoLT, the same finite-layer induction bounds the $L_D$ decoder layers over projected decoder values.
With norm-preserving RoPE, per-head affine RMS QK normalization further bounds attention logits.

At $t>0$ under MoL decoding and across MoL-Hybrid continuation steps, each token still evaluates the feedback map $F_t$ over prefix memory KV rather than running the loop-1 path.
At the same time, 2-loop training under loop forcing supervises only one feedback step from $z^1$ to $z^2$, so it does not train the recursion for extra gains over longer chains.
Together with the three mechanisms above, these properties motivate the empirical evaluation in \cref{sec:execution_results}.
In \cref{tab:decoding_quality_latency}, MoL and MoL-Hybrid decoding achieve similar mean continuation CE to fixed 2-loop decoding on the evaluated windows.
Across 8 windows totaling 7,168 teacher-forced targets, $\Delta$CE relative to fixed 2-loop decoding stays between $-0.0005$ and $+0.0014$.

\subsection{Computation and KV Storage}
\label{sec:cost}
We now show how single-traversal decoding and U-MoLT's cross-only decoder remove repeated per-step traversals, unused decoder traversals, and per-loop prefix KV caches.

\para{Omitting unused decoder traversals}
In U-MoLT, the cross-only decoder reads only encoder-sourced decoder KV and stays outside the encoder feedback path.
\begin{proposition}[Decoder omission]
\label{prop:decoder_omission}
Fix the encoder schedule and its outputs, and consider the deterministic cross-only decoder and prediction head in \cref{eq:decoder,eq:forms}.
Every requested next-token distribution at position $t$ is unchanged when all decoder work is omitted at unrequested loops.
At requested loops, omitting decoder query, cross-attention, and feed-forward updates at other positions also leaves that distribution unchanged, provided decoder KV remain available at $j\leq t$.
\end{proposition}
\begin{proof}
Omitting the decoder at an unrequested loop leaves subsequent encoder loops unchanged because decoder outputs are not fed back.
At any requested position $t$, the decoder starts from the raw encoder output at $t$.
Each decoder layer depends only on the preceding decoder state at $t$ and on decoder KV projected from normalized encoder outputs at $j\leq t$.
Induction over decoder layers makes the next-token distribution at $t$ independent of decoder states at all other positions.
\end{proof}
For next-token inference, the decoder runs its query, cross-attention, and feed-forward updates only at the final prefill position and once per decoding step.
It projects decoder KV only from the final normalized encoder outputs of prefill and of each decoding step.
In multi-loop prefill or decoding, all decoder work is skipped at intermediate loops.
Because \cref{prop:decoder_omission} holds for any fixed encoder schedule, this omission applies to fixed-loop, MoL, and MoL-Hybrid decoding alike.

\para{Repeated block traversals}
Under loop residual, inference remains tied to fixed-loop execution and requires $N$ shared-stack traversals per decoding step, or $NL$ block traversals.
Under loop KV, MoL and MoL-Hybrid decoding instead use one shared-stack traversal per decoding step, compared with $N$ in fixed-loop decoding.
\Cref{tab:execution_costs} compares prefill block traversals, per-step decoding traversals, and prediction targets across U-MoLT schedules.
Loop-forcing training runs the decoder at all $N$ loops.
At inference, \cref{prop:decoder_omission} cuts fixed-loop U-MoLT from $N(L_E+L_D)$ to $NL_E$ full-prompt prefill blocks and from $N(L_E+L_D)$ to $NL_E+L_D$ decoding blocks per step.
For an $L$-layer MoLT-KV stack or a U-MoLT model partitioned with $L_E+L_D=L$ layers, both MoL and MoL-Hybrid decoding execute exactly $L$ block traversals per decoding step, matching a non-looped $L$-layer Transformer.
Relative to fixed 2-loop decoding with decoder omission in U-MoLT, this halves decoding traversals in MoLT-KV from $2L$ to $L$ and reduces U-MoLT from $2L_E+L_D$ to $L_E+L_D$.

Prefill further separates the schedules and architectures.
Empty-cache MoL decoding processes a length-$P$ prompt one token at a time, losing prompt parallelism.
This requires $PL_E$ sequential single-token encoder block traversals in U-MoLT, or $PL$ in MoLT-KV.
MoL-Hybrid-1 and MoL-Hybrid-2 restore parallel prefill with $k\in\{1,2\}$ loops.
An $L$-layer MoLT-KV stack runs $kL$ full-prompt block traversals, doubling non-looped prefill depth at $k=2$.
It then projects prompt memory KV from $z^k_{<P}$ to initialize decoding.
By \cref{prop:decoder_omission}, U-MoLT runs only $kL_E$ full-prompt encoder block traversals.
It then projects prompt memory KV and decoder KV from $z^k_{<P}$ and runs $L_D$ single-token decoder blocks at position $P-1$.
With an equal split $L_E=L_D=L/2$, 2-loop prefill in U-MoLT executes $2L_E=L$ full-prompt block traversals.
MoL-Hybrid-2 therefore matches a non-looped $L$-layer backbone in both full-prompt block depth and per-step decoding traversals.

\begin{table}[htbp]
 \centering
 \caption{\textbf{Prefill and decoding block traversals in U-MoLT.}
 Prefill blocks counts block traversals over a length-$P$ prompt.
 Parallel schedules run each block across all $P$ prompt tokens at once, whereas Mixture-of-Loops runs $P$ sequential single-token steps of $L_E$ encoder blocks.
 In the last three schedules, Prefill blocks counts encoder traversals only.
 With decoder omission, U-MoLT projects prompt decoder KV from the final normalized encoder outputs of prefill and runs $L_D$ single-token decoder blocks only at position $P-1$.
 MoL-Hybrid-1/2 with $k\in\{1,2\}$ prefill loops also projects prompt memory KV from $z^k_{<P}$ after prefill.
 Encoder and Decoder list block traversals per decoding step, where each step consumes one token and predicts the next.
 The first two schedules run $N$ encoder loops with per-loop prefix KV caches.
 Here $m=t-P\geq 0$ indexes continuation tokens, and $s$ indexes the fixed-prompt continuation unroll.
 For MoLT-KV, set $L_E=L$ and $L_D=0$.}
 \label{tab:execution_costs}
 \small
 \setlength{\tabcolsep}{3pt}
 \begin{tabularx}{\textwidth}{@{}llcc>{\raggedright\arraybackslash}X@{}}
 \toprule
 Decoding schedule & Prefill blocks & Encoder & Decoder & Prediction target \\
 \midrule
 Fixed-loop, full decoder & Parallel, $N(L_E+L_D)$ & $NL_E$ & $NL_D$ & Final prediction of the $N$-loop model \\
 Fixed-loop, omitted decoder & Parallel, $NL_E$ & $NL_E$ & $L_D$ & Same final prediction as above \\
 Mixture-of-Loops & Sequential, $P\times L_E$ & $L_E$ & $L_D$ & Stabilized solution, $r\geq t+2$ \\
 MoL-Hybrid-1/2 & Parallel, $kL_E$ & $L_E$ & $L_D$ & Prompt-conditioned continuation, $s\geq m+2$ \\
 \bottomrule
 \end{tabularx}
\end{table}

\para{KV storage}
MoL decoding also reduces persistent KV storage by retaining one prefix history for the shared stack instead of one per loop.
Let $D$ denote the hidden width, $H_{kv}$ the number of KV heads, and $h$ the dimension of each head.
For U-MoLT with MoLA at every encoder layer and one decoder KV history for final-loop predictions, per-layer KV projections give the following payloads for a length-$T$ cached sequence:
\begin{equation}
 C_{\mathrm{Nloop}}=2TH_{kv}h(NL_E+L_D),\qquad
 C_{\mathrm{MoL}}=2TH_{kv}h(L_E+L_D).
 \label{eq:cachepayload}
\end{equation}
MoL-Hybrid decoding has the same payload $C_{\mathrm{MoL}}$ after prefill.
For MoLT-KV, replace $L_E$ with the stack depth $L$ and omit the decoder term $L_D$ in these expressions and the remaining cost analysis.
Even under fixed 2-loop decoding with an equal split $L_E=L_D=L/2$, U-MoLT retains only one decoder KV cache.
Its persistent KV storage therefore stays at $1.5\times$ of a non-looped $L$-layer Transformer rather than $2\times$.
Under MoL and MoL-Hybrid decoding, any U-MoLT partition with $L_E+L_D=L$ and any $L$-layer MoLT-KV stack reach $C_{\mathrm{MoL}}=2TH_{kv}hL$, matching a non-looped $L$-layer Transformer of the same KV width.
By contrast, exact fixed-loop decoding in the full-context Loop Model and MoLT-R baselines retains $N$ full-length sequence KV caches across loops, keeping the $N$-fold KV-cache growth.
These expressions count scalar elements for one sequence, with batch size and bytes per element determining the byte cost.
They count persistent KV caches only and exclude temporary local KV and runtime overhead.

Local KV used only for depth attention can be released after their last reader at the current token.
Sequence and decoder KV instead remain available as prefix caches for later tokens.
In fixed-loop decoding, the shared stack keeps current-loop sequence KV for loop 1 and memory KV for each feedback loop.
In MoL and MoL-Hybrid decoding, only one persistent memory KV cache is retained after prefill.
\Cref{tab:cache} separates these stored objects by lifetime and sharing.

\begin{table}[H]
 \centering
 \caption{\textbf{Memory objects and their lifetimes.}
 Sizes count stored scalar elements for one sequence and exclude batch size and runtime buffers.
 Paired sizes correspond to a length-$T$ parallel loop and a single decoding step, respectively.
 A projected KV cache may coexist with the normalized loop output from which it is computed.
 For MoLT-KV, set $L_E=L$ and omit decoder KV.}
 \label{tab:cache}
 \small
 \setlength{\tabcolsep}{8pt}
 \begin{tabularx}{\textwidth}{@{}ll>{\raggedright\arraybackslash}X@{}}
 \toprule
 Object & Typical size & Lifetime and sharing \\
 \midrule
 Normalized loop output & $TD$ or $D$ & Temporary output used to project memory KV and, in U-MoLT, decoder KV \\
 Sequence KV, per history & $2TL_EH_{kv}h$ & Prefix cache: loop 1 current-loop sequence KV or memory KV \\
 Decoder KV & $2TL_DH_{kv}h$ & U-MoLT: projected per layer from normalized encoder output \\
 Depth KV, full history & $\leq 2TL_EH_{kv}h$ or $\leq 2L_EH_{kv}h$ & Transient local KV retained until the last reader in a length-$T$ loop or decoding step \\
 Depth KV, own only & $2TH_{kv}h$ or $2H_{kv}h$ & Own KV at the active layer, released after attention \\
 \bottomrule
 \end{tabularx}
\end{table}

\para{Projection and attention work}
Let $H_q$ denote the number of query heads and $K_\ell=|\mathcal D_t^{r,\ell}|$ the number of local KV pairs read at layer $\ell$.
Separate memory projections add $2DH_{kv}h$ dense projection parameters per MoLA layer, or $2L_EDH_{kv}h$ for the encoder, shared across loops.
In U-MoLT, decoder KV projections replace standard self-attention KV projections and add no extra parameters.
Keeping MoLA within the $L_E$-layer encoder therefore reduces added parameters and per-step memory projections relative to an $L$-layer MoLT-KV stack.
Reusing local KV projections for depth reads adds no extra projection matrices, while affine QK normalization contributes gain parameters and elementwise work.

The first loop projects current-loop sequence KV at every layer and reuses them for depth attention, requiring $2L_ETDH_{kv}h$ multiply-accumulates over a length-$T$ sequence.
In a parallel feedback loop $r>1$, memory KV projections from $z^{r-1}$ and local KV projections from $X^{r,\ell}$ each require $2L_ETDH_{kv}h$ multiply-accumulates across $T$ positions and $L_E$ layers.
To initialize decoding after $k$-loop prefill over a length-$P$ prompt, MoL-Hybrid decoding additionally projects prompt memory KV from $z^k_{<P}$, requiring $2L_EPDH_{kv}h$ multiply-accumulates.
Each single-token MoL or MoL-Hybrid step reuses cached prefix memory KV without reprojection.
It projects only the new normalized loop output into memory KV and computes current-loop local KV, requiring $2L_EDH_{kv}h$ multiply-accumulates each.
In U-MoLT, decoder KV projections require $2L_DPDH_{kv}h$ multiply-accumulates during prefill and $2L_DDH_{kv}h$ per decoding step.
When $L_E+L_D=L$, local and decoder KV projections match the $L$ standard KV projections per decoding step, leaving only $L_E$ extra memory KV projections.
In equal-split 2-loop prefill, the two encoder loops already run $2L_E=L$ layers of current-loop sequence and local KV projections.
MoL-Hybrid-2 therefore adds loop-2 memory, handoff memory, and prompt decoder KV projections.

Across all $L_E$ MoLA layers, one length-$T$ loop costs $O(L_ET^2H_qh)$ for sequence attention and
\begin{equation}
 O\!\left(TH_qh\sum_{\ell=0}^{L_E-1}K_\ell\right)
 \label{eq:depthcost}
\end{equation}
for depth attention.
For a single decoding step with $T$ cached positions, these costs become $O(L_ETH_qh)$ for sequence attention and $O(H_qh\sum_{\ell=0}^{L_E-1}K_\ell)$ for depth attention.

With full-history depth KV, $K_\ell=\ell+1$ and $\sum_{\ell=0}^{L_E-1}K_\ell=L_E(L_E+1)/2$, giving $O(L_E^2TH_qh)$ per length-$T$ loop and $O(L_E^2H_qh)$ per decoding step.
Because sequence length typically far exceeds encoder depth, full-history depth attention accounts for only an $O(L_E/T)$ fraction of sequence attention in both full-sequence loops and single-token decoding.
This explains why expanding depth access from own KV to full history adds negligible attention FLOPs in \cref{tab:components}.
Keeping MoLA within the $L_E$-layer encoder in U-MoLT reduces this $O(L_E^2)$ term further.
When minimal depth overhead is desired, restricting depth reads to own KV with $K_\ell=1$ lowers depth attention work to $O(L_ETH_qh)$ per length-$T$ loop and $O(L_EH_qh)$ per decoding step.

In U-MoLT, \cref{prop:decoder_omission} also removes quadratic prefill attention in the decoder.
Running the decoder only at position $P-1$ lowers prefill decoder attention from $O(L_DP^2H_qh)$ to $O(L_DPH_qh)$.
Each decoding step adds $O(L_DTH_qh)$ of decoder attention.
Under MoL and MoL-Hybrid decoding, encoder sequence attention and decoder attention over cached positions therefore total $O(LTH_qh)$ when $L_E+L_D=L$.
With an equal split $L_E=L_D=L/2$, 2-loop prefill in U-MoLT matches a non-looped $L$-layer Transformer in quadratic prefill attention, whereas MoL-Hybrid-1 cuts it in half.

Together, these bounds show how U-MoLT and loop KV close the inference efficiency gap in two steps.
First, even under fixed 2-loop execution, \cref{prop:decoder_omission} lets equal-split U-MoLT match non-looped full-prompt block depth and quadratic prefill attention.
It also keeps decoding traversals and persistent KV storage at $1.5\times$ instead of $2\times$, while reaching $3L/2$ blocks of prefix path depth.
Second, MoL and MoL-Hybrid decoding avoid repeating the shared stack at each step and avoid separate per-loop prefix KV caches.
Both architectures therefore run $L$ decoding blocks per step with one persistent prefix KV history for the shared stack.
Under MoL-Hybrid-2, equal-split U-MoLT matches a non-looped $L$-layer Transformer across all four dominant costs while keeping parallel prefill.
Its remaining decoding overhead is small depth attention and $L_E$ extra memory KV projections, while 2-loop prefill adds the decoder readout at $P-1$, prompt decoder KV projections, and extra memory KV projections.

Sparse feedback placement keeps the same persistent KV storage by retaining one current-loop sequence KV cache per non-feedback layer.
Despite the remaining projection, depth-attention, and prompt handoff costs above, MoL-Hybrid decoding lowers total latency relative to fixed 2-loop decoding in the S-size evaluations of \cref{sec:execution_results}.
It cuts total time by 28.0--46.7\% for 128-token prompts with 895 steps in \cref{tab:decoding_quality_latency}, and by 32.1--50.0\% for 4,096-token prompts with 64 steps in \cref{tab:decoding_phase_latency}.
On 4,096-token prompts, U-MoLT with MoL-Hybrid-2 runs in 0.9966\,s versus 1.0006\,s for non-looped OLMo3.

\end{document}